\documentclass[sigconf]{acmart}

\usepackage{times}
\usepackage{epsfig}
\usepackage{graphicx}
\usepackage{amsthm, amsmath}

\usepackage{caption}
\graphicspath{{./figure/}}
\usepackage{multirow}
\usepackage{booktabs}
\usepackage{makecell}
\usepackage{float}
\usepackage{subfigure}
\usepackage{color}

\usepackage{mathrsfs}
\usepackage[noend]{algpseudocode}
\usepackage[switch]{lineno}
\usepackage{epsfig}
\usepackage{lipsum}

\AtBeginDocument{%
  }

\setcopyright{acmcopyright} 
\copyrightyear{2022}
\acmYear{2022}
\acmDOI{3503161.3547951}

\acmConference[MM '22] {Proceedings of the 30th ACM International Conference on Multimedia }{October 10--14, 2022}{Lisbon, Portugal.}
\acmBooktitle{Proceedings of the 30th ACM International Conference on Multimedia (MM '22), October 10--14, 2022, Lisbon, Portugal}
\acmPrice{15.00}
\acmISBN{978-1-4503-9203-7/22/10}

\acmSubmissionID{mmfp0821} 

\begin{document}

\title{Rethinking Super-Resolution as Text-Guided Details Generation}

\author{Chenxi Ma}
\affiliation{
  \institution{School of Computer Science, Shanghai Key Laboratory of Intelligent Information Processing, Shanghai Collaborative Innovation Center of Intelligent Visual Computing, Fudan University}
  \city{}%
  \country{}
}
\email{cxma17@fudan.edu.cn}

\author{Bo Yan}
\authornote{Corresponding Author. This work is supported by NSFC (Grant No.: U2001209, 61902076) and Natural Science Foundation of Shanghai (21ZR1406600, 21ZR1403300).}
\affiliation{
  \institution{School of Computer Science, Shanghai Key Laboratory of Intelligent Information Processing, Shanghai Collaborative Innovation Center of Intelligent Visual Computing, Fudan University}
  \city{}%
  \country{}
}
\email{byan@fudan.edu.cn}

\author{Qing Lin}
\affiliation{
  \institution{School of Computer Science, Shanghai Key Laboratory of Intelligent Information Processing, Shanghai Collaborative Innovation Center of Intelligent Visual Computing, Fudan University}
  \city{}%
  \country{}
}
\email{18210240028@fudan.edu.cn}

\author{Weimin Tan}
\affiliation{
  \institution{School of Computer Science, Shanghai Key Laboratory of Intelligent Information Processing, Shanghai Collaborative Innovation Center of Intelligent Visual Computing, Fudan University}
  \city{}%
  \country{}
}
\email{wmtan14@fudan.edu.cn}

\author{Siming Chen}
\affiliation{
  \institution{School of Data Science, Fudan University}
  \city{}%
  \country{}
}
\email{simingchen3@gmail.com}

\renewcommand{\shortauthors}{Chenxi Ma, Bo Yan, Qing Lin, Weimin Tan, Siming Chen}

\begin{abstract}
  Deep neural networks have greatly promoted the performance of single image super-resolution (SISR). Conventional methods still resort to restoring the single high-resolution (HR) solution only based on the input of image modality. However, the image-level information is insufficient to predict adequate details and photo-realistic visual quality facing large upscaling factors ($\times$8, $\times$16). In this paper, we propose a new perspective that regards the SISR as a semantic image detail enhancement problem to generate semantically reasonable HR image that are faithful to the ground truth. To enhance the semantic accuracy and the visual quality of the reconstructed image, we explore the multi-modal fusion learning in SISR by proposing a \textbf{T}ext-\textbf{G}uided \textbf{S}uper-\textbf{R}esolution (\textbf{TGSR}) framework, which can effectively utilize the information from the text and image modalities. Different from existing methods, the proposed TGSR could generate HR image details that match the text descriptions through a coarse-to-fine process. Extensive experiments and ablation studies demonstrate the effect of the TGSR, which exploits the text reference to recover realistic images.
\end{abstract}

\ccsdesc[300]{Computing methodologies~Reconstruction}

\keywords{single image super-resolution, text-guided super-resolution, multi-modal fusion learning}

\maketitle

\begin{figure}[!h]
\centering  \includegraphics[width=0.49\textwidth]{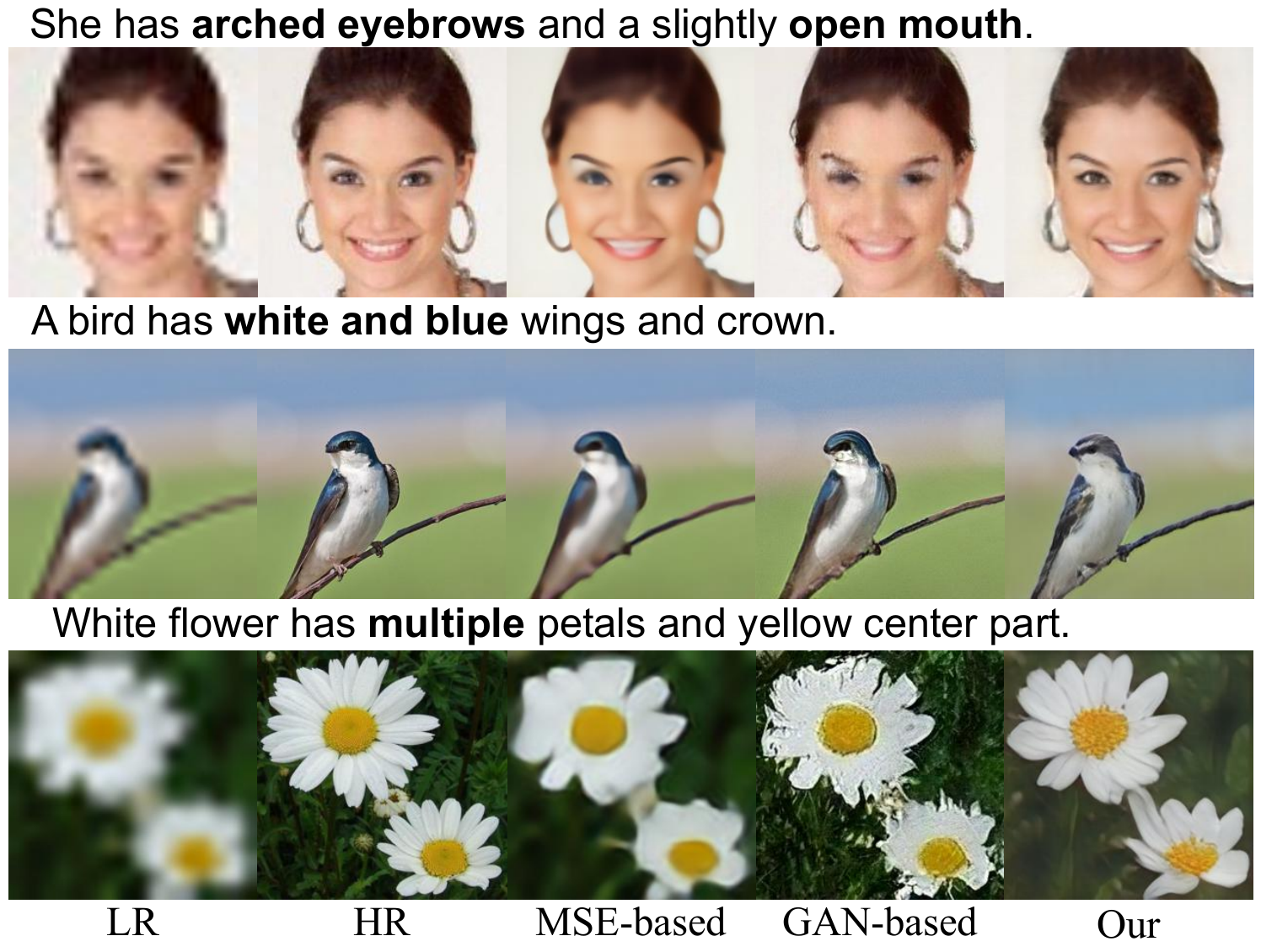}
\vspace{-0.8em} \caption{The large factor SR results of MSE-based SISR (DRN~\cite{drn}) and face SR methods (SuperFAN~\cite{superfan}), GAN-based SISR (SPSR~\cite{spsr}) and face SR (DICGAN~\cite{dicgan}) methods, and our models. Our model can make better use of text descriptions to restore clear and rational image details.}
\label{f0}  \end{figure}

\section{Introduction}

Since the LR image is too small to contain enough information for large-factor SR. To further enhance the SR performance, some external priors are introduced to provide more guidance for SR models. FSRNet~\cite{fsrnet}, DeepSEE~\cite{deepsee}, and Christian et al.~\cite{shen} exploit the face structure prior (face parsing map, face landmark heatmap) to restore face images. The audio-aided face SR~\cite{audioSR} method utilizes the audio prior to guide the face super-resolution problem by extracting facial attributes (age, gender, ethnicity) from the voice of a speaker. These priors are only helpful for face images and can not generalize to natural images. SFTGAN~\cite{sftgan} utilizes semantic segmentation maps to super-resolve LR images. However, these image-level semantic priors contain limited information and require additional calculation of existing semantic extraction methods, the accuracy of which greatly affects the super-resolution performance.

In comparison to the image-level prior, text description of an image contains more abundant semantic information, and describes global image style and local features of main objects, such as, color, shape, species, age, emotion, etc. The text description is easily available, and it can intuitively and flexibly express concepts of an image, it is helpful for us to picture the rough image contents in our mind. By utilizing text information in SR, we can increase the flexibility and controllability of SR, so that we can better meet expectations of people. For old photo recovery, the color or other details are contaminated, additional text information is useful to control the image contents and ensure the SR result conform to the common sense. In addition, for the surveillance video, investigators often draw a profile of characters according to descriptions of witnesses and low-quality surveillance video frames to search for suspects. This requires the SR method to generate face attributes that conform to descriptions, in addition to clear and reasonable textures. However, it is difficult for previous SR methods to restore HR images with specific attributes.

To address above issues, we first reveal the potential of text descriptions in SISR, that the text can provide important reference for SR approaches to restore reasonable results. Here, we propose a \textbf{T}ext-\textbf{G}uided \textbf{S}uper-\textbf{R}esolution (\textbf{TGSR}) method, which uses the multimodal fusion learning to integrate text semantic information into large-factor SR.
Since text and image are data of totally different modalities, how to effectively fuse the two modalities and understand semantic information from text descriptions for better SR is a challenging task. The proposed TGSR adopts a coarse-to-fine framework to restore image details. The text information is embedded in the coarse SR stage to enhance image features and restore a rough SR image. In the fine SR stage, the TGSR further refines the visual quality of the final result. We also improve semantic accuracy of the SR result based on the text description during training. Therefore, the text description can guide the SR model to generate more accurate details and meanwhile manipulate the color, shape, texture, background, and other image characteristics.

As shown in Figure~\ref{f0}, if the LR image is too small (large-factor SR) and lack some visual information, traditional SR methods (DRN~\cite{drn}, SuperFAN~\cite{superfan}, SPSR~\cite{spsr}, DICGAN~\cite{dicgan}) produce blur and fake artifacts due to lacking the understanding of the contextual knowledge. In comparison, the proposed TGSR can effectively understand text descriptions and restore photo-realistic and reasonable results. In addition, the TGSR can generate diverse valid HR solutions for a single LR input by manipulating text descriptions, which enhances the flexibility and the practicability of SR. 

The main contributions are as follows:

\begin{itemize}
\item This paper rethinks the SR as a semantic detail enhancement problem, that restores semantically reasonable HR details.
\item This paper proposes the first text-guided image super-resolution (TGSR) approach based on the multimodal fusion learning, and explores the effectiveness of text descriptions to large-factor SR.
\item The proposed TGSR adopts a coarse-to-fine process to restore an HR image, and introduces the text features through the text attention module to modulate image features. The text-guided losses constrain the network to pay more attention on image regions focused by text and to generate semantically reasonable textures. 
\end{itemize}

\section{Related Work}
\subsection{Single Image Super-Resolution}
\textbf{Deep Learning Based Super-Resolution:}
Deep learning based methods have dominated the development of SISR for their strong representation and fitting capabilities. SRCNN~\cite{srcnn}, composed of three convolution layers, first introduces the convolutional neural network (CNN) into single image super-resolution (SISR) and leads to a dramatic leap. Based on this work, following SISR methods achieve continuous breakthroughs by proposing
different network structures (VDSR~\cite{vdsr}, EDSR~\cite{edsr}, DRN~\cite{drn}, Liu et al.~\cite{rfa}).
By exploiting the generative adversarial network (GAN)~\cite{gan} in SISR, SRGAN~\cite{srgan}, ESRGAN~\cite{esrgan}, SPSR~\cite{spsr} generates visually pleasing results with more high-frequency details. 

To further improve the SR performance, some recent methods (FSRNet~\cite{fsrnet}, DeepSEE~\cite{deepsee}, SFTGAN~\cite{sftgan}) introduce external priors (facial landmark heatmaps, parsing maps, semantic maps) to provide guidance for face image SR. To, SFTGAN~\cite{sftgan} adopts the semantic segmentation result of a LR image to utilize the semantic prior for SR of natural images. 

\textbf{Explorative Super-Resolution: }
Most traditional SISR methods focus on outputting a unique HR image to approximate the ground truth and ignore the abundance of plausible HR explanations to the input LR image. Due to the ill-posed nature of the SR task, several recent works~\cite{explorable, pulse} are proposed to break this limitation and explore infinitely many plausible reconstructions for a given LR image. Explorable Super-Resolution~\cite{explorable} generalizes the traditional SISR task toward an image restoration task, that can output different possible HR images with the observed LR image and can also support editing the output image through user interaction. In specific, the explorable SR method comprises a graphical user interface with a SR network.

PULSE~\cite{pulse} uses the GAN prior to generates a HR face image by optimizing the latent vector of a pre-trained GAN. The DeepSEE~\cite{deepsee} also achieves explorative face super-resolution by controlling the semantic maps and the style vector to change the shape and appearance of specific semantic regions. The audio-aided face SR method~\cite{audioSR} introduces the audio prior to guide the face SR by exploiting an encoder-decoder to fuse features from the voice and image of a speaker. In this way, it is able to reconstruct different HR face images that have consistent characteristics with different input audios. These explorative SR works reveal more possibilities for image super-resolution. 

\subsection{Text Guided Image Reconstruction}
Though the text prior has never been explored in SR, it is utilized in several computer vision fields, such as text-to-image synthesis, text-guided image colorization, these ideas inspire us to explore the SR task with text information, which is first done in the SR field.

\textbf{Text-to-Image Synthesis:} 
Generating images from text descriptions, a classical multimodal task, has received a lot of attention in academia. \cite{reed} first used a GAN conditioned on text features to synthesis images. AttnGAN~\cite{attngan} proposes an attentional generative network and a deep attentional multimodal similarity model to synthesize fine-grained details at different sub-regions of the image by paying attention to the relevant words in the image description. StackGAN~\cite{stackgan} generates images in two stages. The first stage focuses on the rough background, color, and contour components and the second stage focuses on image details. In addition to improving the generation quality, MA-GAN \cite{msaan} guarantees the generation similarity of related sentences describing the same image by proposing a Single-sentence Generation and Multi-sentence Discrimination (SGMD) module. XMC-GAN~\cite{cmcl} uses an attentional self-modulation generator and a contrastive discriminator to enforce text-image correspondence. ManiGAN~\cite{manigan} proposes a text-image affine combination module to select and correlate image regions relevant to given text and a detail correction module to rectify mismatched attributes and completes missing contents. 

\textbf{Text-guided Image Colorization:}
Image colorization aims to generate color of a gray-scale image. To utilize text descriptions to edit the image color, LBIE~\cite{lbieram} proposes a recurrent attentive model to fuse image and language features. Text2Colors~\cite{Text2Colors} adopts a text-to-palette generation network and a palette-based colorization network to capture semantic information and produce relevant color palettes. Tag2Pix~\cite{Tag2Pix} utilizes the text tag related to the color, and proposes a line art colorization method.

\section{Text-Guided Image Super-Resolution (TGSR)}
This section first analyses the effect of text on SISR, proposes the multimodal fusion network for text-guided SISR, and introduces the training constraints.

\subsection{Problem Formulation}
Given a LR image as well as its text description, the goal of the text-guided super-resolution is reconstructing a HR image with rational textures. Different from the traditional SR, the key to TGSR is reasonably utilizing the text features and giving the SR network more guidance. To make better use of the text guidance, we need to solve two problems: what kind of information text can provide and how such information works.

The text information can be the global description or the object feature of an image. First, text can indicate salient objects in an image and guide the SR network to find these objects and focus on important image regions. Therefore, we introduce a Text Attention Module to exploit the response of keywords in text to corresponding image regions. Second, text describes visual features (color, shape, etc.) and semantic features (species, gender, age, etc.) of an image. Therefore, we adopt the text-image consistency loss to enforce the semantic consistency between SR results and text descriptions. Thus, the text is able to tell the SR network the significant image feature and is helpful for the network to restore more real textures and details by utilizing the visual relationship between text and image. which ensures the abundance of details

\begin{figure*}[tb] \centering \includegraphics[width=0.9\textwidth]{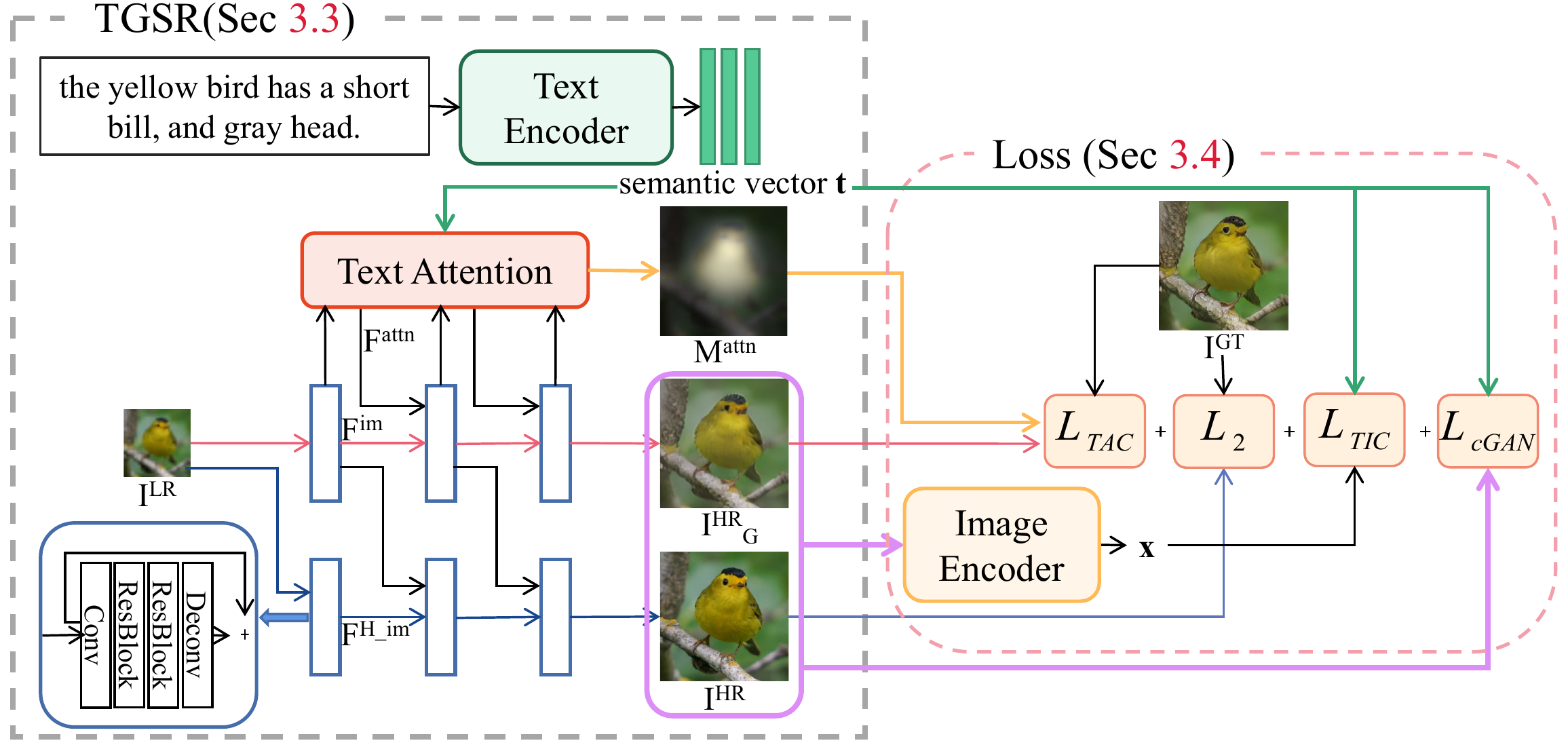} \vspace{-0.1em} 
\caption{Architecture of the text-guided super-resolution (TGSR) framework, composed of a global branch and a refine branch. The pre-trained text encoder first extracts a semantic vector $t$ from the text description. The global branch generates a rough HR image $I^{HR}_{G}$ by incorporating $t$ and image features with a text attention module. The refine branch restores more high-frequency details of the final HR image $I^{HR}$. The total loss of the TGSR is composed of the L2 loss, the text-attention reconstruction loss, the text-image consistency loss and the conditional GAN loss.} \label{fstruct} \end{figure*}

\subsection{Overview}
In Figure~\ref{fstruct}, the TGSR adopts a coarse-to-fine process through a dual-branch network structure to restore more precise image details. The text features are extracted and embedded in the coarse SR stage to obtain a rough SR result consistent with the text description. The fine SR stage takes the output of the first stage as input to refine the final SR result.

Suppose $(I, text)$ represents an image-text pair, the text feature $t$ is extracted from the text description $text$ of the LR image $I^{LR}$. Then, the TGSR network, composed of a text-aware global branch and a refine branch, utilizes the external text guidance and the internal image information from $t$ and $I^{LR}$ to restore visually compromising HR images. The text-aware global branch uses the text guidance to generate a rough SR result $I^{HR}_{G}$ by incorporating the text feature $t$ with image features. Based on the $I^{HR}_{G}$, the refine branch focuses on refining photo-realistic details and increases the fidelity of the final HR result $I^{HR}$. The TGSR is optimized under constraints at image and semantic levels.

\subsection{Multi-Modal Fusion SR Network}

\textbf{Text Encoder:}
\label{sectife}
The text encoder chooses the bi-directional Long Short-Term Memory network (LSTM)~\cite{lstm} to extract the text feature, inspired by the AttnGAN~\cite{attngan}. In specific, the text description is transferred to the text feature extractor to extract a preliminary text feature $t \in R^{D*T}$, where T is the number of words and D is the dimension of each word vector in the text feature. 
\begin{equation}
\begin{array}{lr}
t = TextEncoder(text).
\end{array}
\label{eq1}
\end{equation}

\textbf{SR Network:}
As shown in Figure~\ref{fstruct}, the dual-branch TGSR network contains a global branch and a refine branch to generate a coarse SR result and enrich image details, respectively. Both the global branch and the refine branch adopt a pyramid structure to reconstruct $\times$2, $\times$4, $\times$8 upscaled image features. 

In the global branch, we extract and upscale feature ($F^{im}_{1}$) from the input LR image with a convolution layer, a residual block, and a deconvolution layer, which are wrapped together by the blue block in Figure~\ref{fstruct}. The residual block is composed of two convolution layers and a residual connection.

\begin{figure}[tb]
\centering \includegraphics[width=0.48\textwidth]{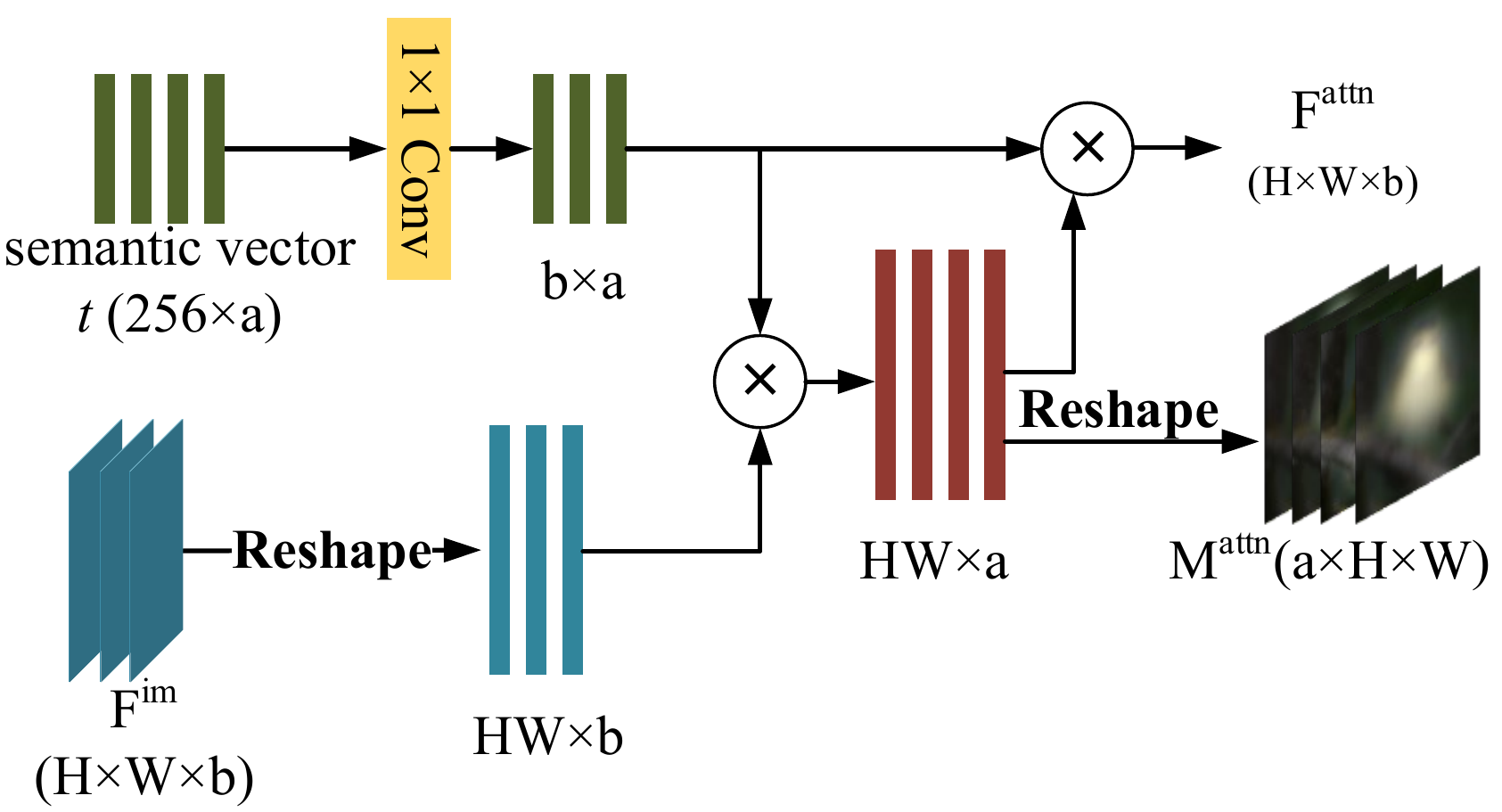} \vspace{-0.1em}
\caption{Structure of the text attention module (TAM). The TAM generates a text-embedding image feature $F^{attn}$ and an attention map $M^{attn} \in R^{a \times H \times W}$ corresponding to $a$ words in text.}
\label{ftam} \end{figure}

\begin{figure}[tb]
\leftline{This bird has gray upper body and white breast.}
\centering
 \begin{minipage}[b]{0.48\textwidth}\centering \includegraphics[width=1\textwidth]{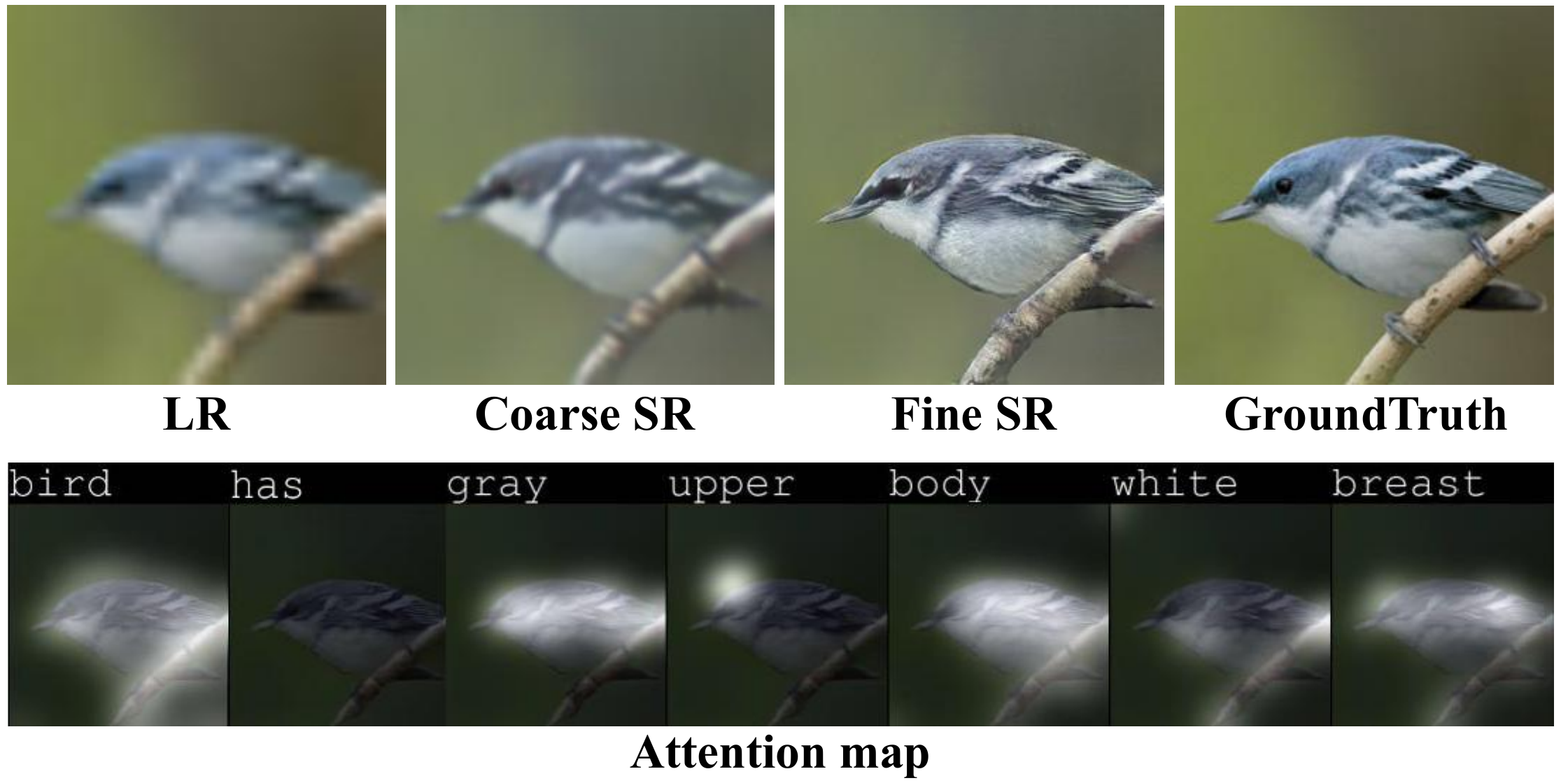} \end{minipage} \vspace{-0.8em}
\caption{The outputs of the global branch and the refine branch and the word attention map corresponding to each word in the input text description.}
\label{fHL}  \end{figure}

Then, the text attention module (TAM)~\cite{attngan}, shown in Figure~\ref{ftam}, is adopted to combine the text feature and the image feature. The TAM calculates a word attention map $M^{attn}_{i}$ and a text-embedding image feature $F^{attn}_{i}$ based on the text feature $t$ and $F^{im}_{i}$ at $i$-th stage. With the explainable text attention maps $M^{attn}$ from the TAM (see Figure~\ref{fHL}), the text feature can accurately focus on different image regions and is helpful for the TGSR to well understand the key semantic information.
\begin{equation}
 \begin{array}{lr}
 M^{attn}_{i}, F^{attn}_{i} = TAM(t, F^{im}_{i}).
\end{array}
\label{eq2}
\end{equation}

Next, two image features $F^{attn}_{i}$ and $F^{im}_{i}$ are concatenated and delivered to a convolution layer, a residual block, and a deconvolution layer to obtain the image feature $F^{im}_{i+1}$ at next stage. By progressively incorporating text features and image features, we can fully utilize the text guidance. Last, one convolution layer with kernel size 3$\times$3$\times$3 is operated on $F^{im}_{i}$ to obtain a coarse HR images ($I^{HR}_{G}$). 

The refine branch takes the LR image $I^{LR}$ and the image feature $F^{im}_{i}$ at different scales of the global branch as input to focus on refining authentic image details. Similar to the global branch, the refine branch extracts the feature of $I^{LR}$ and progressively upscales it to obtain $F^{H\_im}_{i}$.

Then, the feature $F^{H\_im}_{i}$ at each stage are cascaded with $F^{im}_{i}$. The final HR image ($I^{HR}$) is generated through a convolution layer with kernel size 3$\times$3 and 3 filters. As shown in Figure~\ref{fHL}, the fine SR result of the refine branch is able to enhance the image details and restore more faithful textures based on the coarse SR result of the global branch.

The convolution layers, used to output images, have kernel size 3 and output channel number 3. Other convolution layers are followed by a Rectified Linear Unit (ReLU) activation layer and have kernel of size 3$\times$3 and 64 filters. Each deconvolution layer has kernel of size 6$\times$6 with stride 2 and 64 filters.

\subsection{Training Constraints}
The goal of our TGSR is simultaneously restoring accurate image details and rational image contents, that are consistent with the semantic information of the text prior. Since it is difficult to optimize the SR network at pixel level and semantic level directly, we separate the final optimization objective into two stages. To enable the network in the coarse SR stage to focus on rough image structure and high-level semantic accuracy, the low-frequency component $\hat{I}^{GT}$ of the ground-truth HR image $I^{GT}$ serves as the training label of the global branch, which has complete semantic information but few high-frequency textures. In this way, the refine branch, which is supervised by $I^{GT}$, has less training difficulty and can largely improve the pixel accuracy. The $\hat{I}^{GT}$ is obtained by applying the low-pass filter on $I^{GT}$.

The TGSR is trained in an end-to-end way. Suppose $(I_{n}, text_{n})_{n=1:M}$ denote a batch of image-text pairs. The global branch and the refine branch are jointly optimized with a global constraint and a fine constraint ($\mathcal{L}_{global}, \mathcal{L}_{fine}$).
\begin{equation}
\mathcal{L} = \mathcal{L}_{global} + \mathcal{L}_{fine}.
\label{eq3-1}
\end{equation}

As shown in Figure~\ref{fstruct}, the $\mathcal{L}_{global}$, calculated on the $I^{HR}_G$ of the global branch, contains a L2 reconstruction loss, a text-image consistency loss $\mathcal{L}_{TIC}$ (DAMSM loss~\cite{attngan}), and a text-adaptive conditional GAN loss $\mathcal{L}_{cGAN}$. The $\mathcal{L}_{fine}$, calculated on the output $I^{HR}$ of the refine branch, is composed of a text-attention reconstruction loss $\mathcal{L}_{TAR}$ in addition to $\mathcal{L}_{cGAN}$ and $\mathcal{L}_{TIC}$.

\begin{equation}  \begin{split} \vspace{1.6ex}
\mathcal{L}_{global} = & \lambda_{L2}||I^{HR}_{G}, \hat{I}^{GT}||_{2} + \lambda_{cGAN}\mathcal{L}_{cGAN}(I^{HR}_{G}, t) \\
 & + \lambda_{TIC}\mathcal{L}_{TIC}(I^{HR}_{G}, t), \\
\mathcal{L}_{fine} = & \lambda_{TAR}\mathcal{L}_{TAR}(I^{HR}, I^{GT}) + \lambda_{cGAN}\mathcal{L}_{cGAN}(I^{HR}, t) \\ & + \lambda_{TIC}\mathcal{L}_{TIC}(I^{HR}, t),
\end{split}  \label{eq42}  \end{equation}
where $\lambda_{L2}$, $\lambda_{cGAN}$, $\lambda_{TIC}$, and $\lambda_{TAR}$ represent the weights used to balance contributions of different losses.

\textbf{The text-adaptive conditional GAN loss:} $\mathcal{L}_{cGAN}$ utilizes the conditional GAN, where the conditional discriminator $cD(\cdot)$ judges the realness of image $I (I^{HR}, I^{HR}_{G})$ conditioned on the text feature $t$.
 \begin{equation}
\mathcal{L}_{cGAN}(I, t) = \mathcal{E}[\log(cD(I|t))].
\label{eq5}  \end{equation}

\textbf{The text-image consistency loss:} $\mathcal{L}_{TIC}$ constrains semantic consistency between an image and its text description. AttnGAN~\cite{attngan} proposes the deep attentional multimodal similarity module to map image and text into a common space and to calculate similarity between them. We introduce this module into our TGSR network to calculate the text-image matching score (TIM) and the $\mathcal{L}_{TIC}$.

As described before, the text encoder maps a text description into the semantic vector $t$. Then, we map the super-resolved image $I (I^{HR}, I^{HR}_{G})$ into the image feature $x\in 768\times17\times17$ with an image feature extractor (Inception-v3~\cite{inceptionv3}). The $x$ is reshaped to $R^{768\times289}$, and $x_{i} \in R^{768}$ denotes the visual feature vector for $i_{th}$ image region.

To evaluate the semantic accuracy of the SR results, the TIM measures the matching degree between an image and its text description by the following calculations:

First, we generate and normalize the similarity matrix, $s\in R^{T\times289}$, where $s_{i,j}$ is similarity between $i_{th}$ word and $j_{th}$ image sub-region).
\begin{equation}  \begin{array}{lr}\vspace{1.6ex}
s=t^{T}x,\\
s_{i,j}'=exp(s_{i,j})/\sum_{k=0:T-1}exp(s_{k,j}).
\end{array} \label{eq6} \end{equation}

Then, we obtain the region-context vector $c$, where $c_{i}$ denotes the relation between $i_{th}$ word and image subregions.
\begin{equation}  \begin{array}{lr} \vspace{1.6ex}
a_{j}=exp(s_{i,j}')/\sum_{k=0:288}exp(s_{i,k}'),\\
c_{i}=\sum_{j=0:288}a_{j}x_{j}.
\end{array} \label{eq6} \end{equation}

Last, the relevance $R(c_{i},t_{i})$ between $i_{th}$ word and image is generated to obtain the $TIM(I_{n},text_{n})$.
\begin{equation}  \begin{array}{lr} \vspace{1.6ex}
R(c_{i},t_{i})=(c_{i}^{T}t_{i})/(||c_{i}||\cdot ||t_{i}||),\\
TIM(I_{n},text_{n})=log(\sum_{i=1:T-1}exp(\cdot R(c_{i},t_{i}))).
\end{array} \label{eq6} \end{equation}

The $L_{TIC}$ is the negative log posterior probability between the image and text.
\begin{equation}  \begin{array}{lr}\vspace{1.6ex}
P(text_{n}|I_{n})=\frac {exp(R(c_{n_i},t_{n_i}))}{\sum_{j=1:M}exp(R(c_{n_j},t_{n_j}))};\\
L_{TIC}=-(\sum_{n=1:M}\log P(text_{n}|I_{n})\\ +\sum_{n=1:M}\log P(I_{n}|text_{n})).
\end{array} \label{eq6} \end{equation}

\textbf{The text-attention reconstruction loss:} $\mathcal{L}_{TAR}$ can be considered as a weighted L2 loss, which calculates the pixel-level similarity between the ground truth $I^{GT}$ and the super-resolved image $I^{HR}$. The $\mathcal{L}_{TAR}$ assigns pixels different weights based on five attention maps with the largest activation values in $M^{attn}$, that corresponding to attention maps of five keywords, to make the network focus on the accuracy of pixels in visually important regions.
\begin{equation}
\mathcal{L}_{TAR} = \frac{1}{5}\sum_{i=1:5}M^{attn}_{i}||I^{HR}, I^{GT}||_{2}.
\label{eq7}  \end{equation}

The text feature encoder and the image feature encoder are pretrained following the methods~\cite{attngan} and keep fixed when training the TGSR model. 

\section{Experiments and Analysis}
In this section, we first introduce the datasets and implementation details. Then, we conduct the ablation study to evaluate the contributions of different designs and compare the proposed TGSR with state-of-the-art SR methods. At last, the diversity of SR results is demonstrated by manipulating the text descriptions. 

\begin{table}[tb]
\centering \scalebox{0.7}[0.7]{
\begin{tabular}{|c|c|c|c|c|c|c|c|c|}
\hline
\multirow{2}{*}{dataset} & \multicolumn{2}{c|}{CUB}& \multicolumn{2}{c|}{Oxford-102}& \multicolumn{2}{c|}{CelebA}& \multicolumn{2}{c|}{COCO}\\
\cline{2-9}
 &Train& Test& Train& Test& Train& Test&Train&Test\\
\hline
Images&8855&2933&7,034&1,155&162,770&39,829&80,000&40,000 \\
\hline
Text per image&10&10&10&10&10&10&5&5 \\
\hline
\end{tabular}  } 
\caption{The datasets statistics.} \label{tdata}  \end{table}

\subsection{Datasets and Protocols}
We train and test our models on Caltech-UCSD Birds 200 (CUB)~\cite{cub}, Oxford-102~\cite{flower}, CelebA~\cite{celeba} and COCO~\cite{coco} datasets, where all images are annotated with several natural language captions. The details of datasets are listed in Table~\ref{tdata}. All images are resized and cropped into patches of size 256. To train SR models, LR images are obtained by downscaling HR images with bicubic interpolation.

All models are trained on the machine with 2.20 GHz Intel (R) Xeon (R) CPU, and GTX1080Ti GPU (128G RAM). The initial learning rate is set to 1e-4. We adopt Adam optimizer with $\beta$1 = 0.9, $\beta$2 = 0.999, $\varepsilon$ = 1$e$-8. The loss weights $\lambda_{L2}$, $\lambda_{cGAN}$, $\lambda_{TIC}$, and $\lambda_{TAR}$ are set as 1, 0.1, 0.5, and 1 respectively.

The PSNR and SSIM~\cite{ssim} measure the distortion degree of images and ignore the subjective quality. As stated in~\cite{pipal}, NIQE~\cite{niqe} and Perceptual Index (PI) (the lower, the better) cannot fairly reflect the subjective performance, since they cannot distinguish the GAN generated noises and real details and prefer images with obvious unrealistic artifacts produced by GAN-based methods. Since our TGSR aims to output better visual results with accurate semantic features, we choose the R-precision (TIM), which is used to measure the consistency between images and text descriptions, by following~\cite{attngan,textimgan} to evaluate the recovery degree of the semantic information in the super-resolved image according to the text. 
\begin{figure}[tb]
\centering  \includegraphics[width=0.49\textwidth]{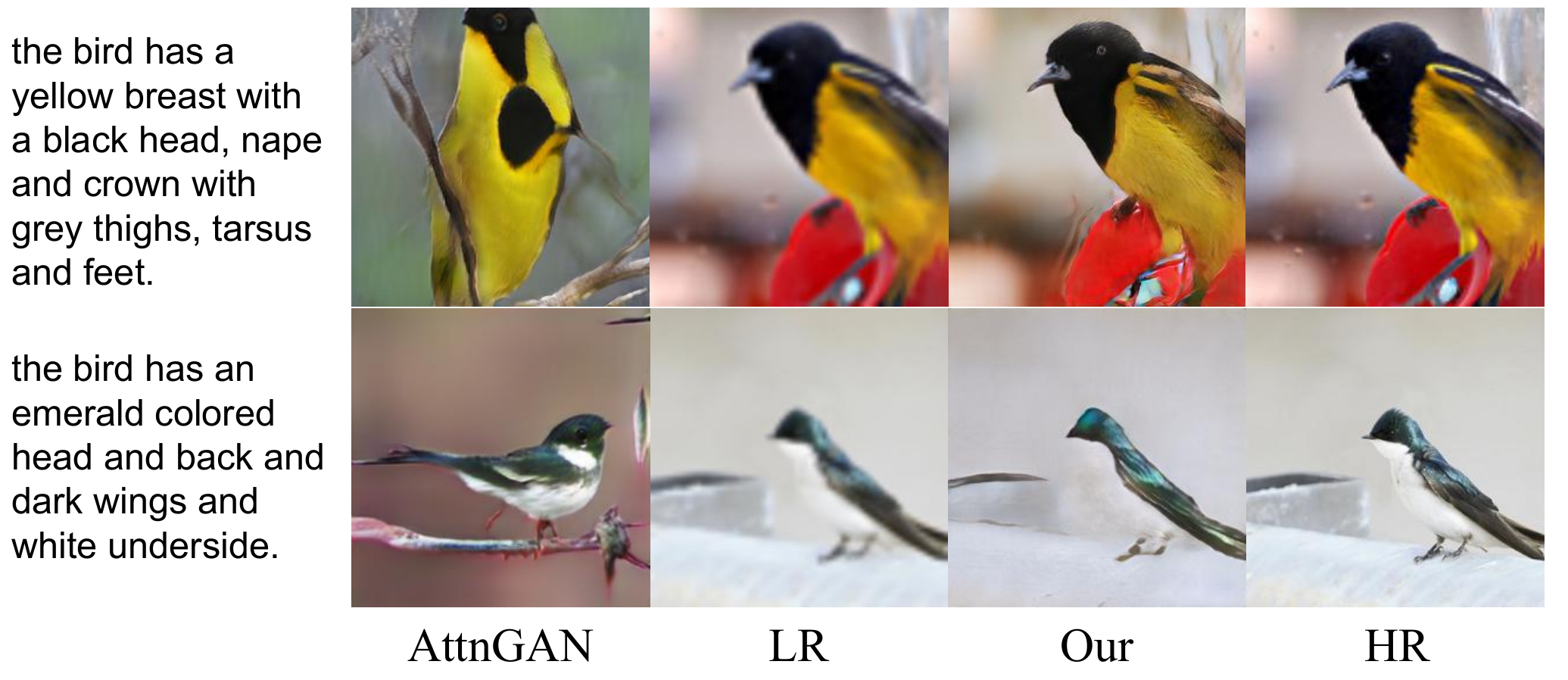} \vspace{-0.3em}
\caption{A visual comparison to the text-to-image synthesis model AttnGAN~\cite{attngan}.}
\label{fattn}  \end{figure}

\subsection{Ablation Study}
To verifies the effect of text to SR, we first evaluate models with different inputs, including text, LR image, and the combination of text and LR image. Then, we construct different models to illustrate contributions of different designs.
\begin{table}[tb]
\centering \scalebox{0.82}[0.82]{
\begin{tabular}{c|cccccc}
\toprule     \toprule
&baseline& + TAM& + $\mathcal{L}_{TIC}$& + coarse-to-fine& + $\mathcal{L}_{TAR}$\\ 
\midrule
NIQE$\downarrow$&10.801&4.657 &4.650&3.424 &\textbf{3.179}\\ 
PI$\downarrow$&9.048 &3.513&3.840&3.239&\textbf{2.663}\\
TIM$\uparrow$&1.453&2.643&\textbf{3.641}&3.501&3.553\\
\bottomrule    \end{tabular} } 
\caption{Experimental results of ablation study. The average NIQE/PI/TIM on CUB for $\times$ 8 SR. TIM shows the semantic accuracy of images. $\downarrow$ denotes the lower is the better.}
\label{tabnet}   \end{table}

\begin{figure}[tb]
\leftline{\small{A small bird with a \textbf{yellow} head, a light brown and yellow throat.}}
\begin{minipage}[t]{1.3cm} \centering  \includegraphics[width=1.4cm]{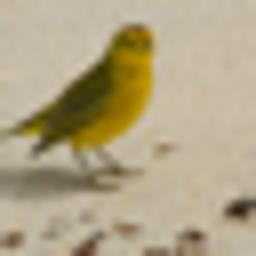}  \\ \centering{\small{Bicubic}} \   \end{minipage}
    \begin{minipage}[t]{1.3cm}  \centering \includegraphics[width=1.4cm]{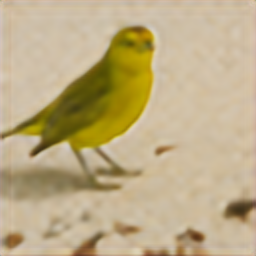} \\ \centering{\small{Baseline}} \  \end{minipage}
    \begin{minipage}[t]{1.3cm} \centering  \includegraphics[width=1.4cm]{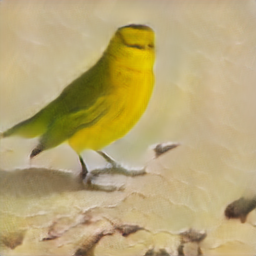}  \\ \centering{\small{+ TAM}}  \ \end{minipage}
    \begin{minipage}[t]{1.3cm}  \centering  \includegraphics[width=1.4cm]{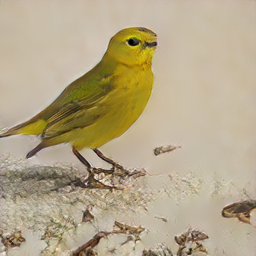} \\ \centering{\small{+ $\mathcal{L}_{TIC}$}} \ \end{minipage}
   \begin{minipage}[t]{1.3cm}  \centering  \includegraphics[width=1.4cm]{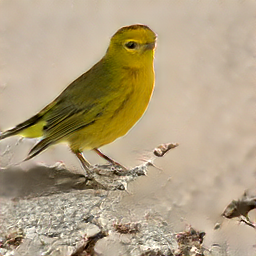}  \\ \centering{\small{+ coarse-to-fine}}  \ \end{minipage}
    \begin{minipage}[t]{1.3cm}  \centering \includegraphics[width=1.4cm]{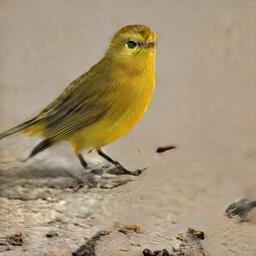} \\ \centering{\small{+
    $\mathcal{L}_{TAR}$}}  \ \end{minipage}
\leftline{\small{A small bird with a \textbf{red} head, a light brown and yellow throat.}}
  \begin{minipage}[t]{1.3cm} \centering  \includegraphics[width=1.4cm]{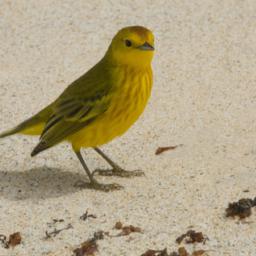} \\ \centering{\small{GT}} \
    \end{minipage}
  \begin{minipage}[t]{1.3cm}  \centering \includegraphics[width=1.4cm]{fattenMSE1.png} \\ \centering{\small{Baseline}}    \end{minipage}
  \begin{minipage}[t]{1.3cm}  \centering \includegraphics[width=1.4cm]{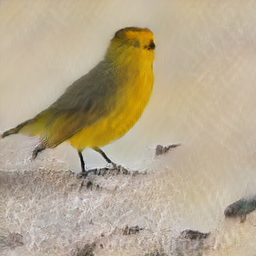} \\ \centering{\small{+ TAM}}   \end{minipage}
  \begin{minipage}[t]{1.3cm} \centering  \includegraphics[width=1.4cm]{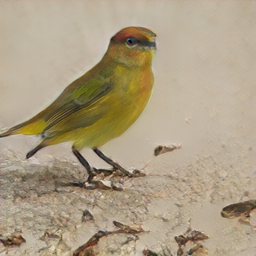} \\ \centering{\small{+
    $\mathcal{L}_{TIC}$}} \ \end{minipage}
  \begin{minipage}[t]{1.3cm}  \centering  \includegraphics[width=1.4cm]{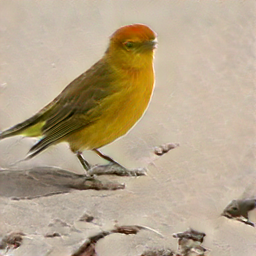} \\ \centering{\small{+ coarse-to-fine}} \end{minipage}
 \begin{minipage}[t]{1.3cm}  \centering \includegraphics[width=1.4cm]{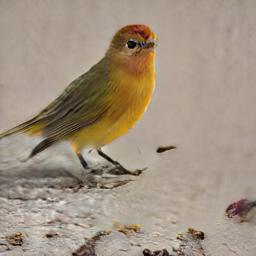} \\ \centering{\small{+    $\mathcal{L}_{TAR}$}} \end{minipage}  \vspace{-0.5em}
\caption{A visual comparison of ablation study. The upper and lower lines denote two SR results of different models on one LR image. Different results are obtained by inputting two texts with few changes to the same model.} \label{fabnet}   \end{figure}

\textbf{Text only.} Figure~\ref{fattn} compares our model with the text-to-image synthesis model, AttnGAN~\cite{attngan}. The AttnGAN uses the same text descriptions as ours to generate images, as we can observe that the output image of the AttnGAN has accurate semantic features but lacks pixel-wise accuracy.

\textbf{LR to HR.} To verify the contribution of text to SR, we first remove the text input from our model and construct a baseline, which is similar to traditional SISR models and only requires LR images as input, by cascading several convolution layers, residual blocks and deconvolution layers. The baseline model is trained with the L2 loss only. As shown in Table~\ref{tabnet}, the NIQE/PI/TIM scores of outputs of the baseline model are lower than others. In addition, as shown in Figure~\ref{fabnet}, the baseline can not use the external text information and only outputs a single SR result for a LR image input.

\textbf{Text and LR to HR.} Then, we introduce the text input into the baseline SR model with the proposed TAM modules. Other ablation studies are also conducted by constructing models with different configurations to analyze the effectiveness of different designs of the TGSR, including the coarse-to-fine structure, the $\mathcal{L}_{TIC}$ and $\mathcal{L}_{TAR}$, by progressively added them on the baseline. 

Figure~\ref{fabnet} denotes different SR results of one LR image with different text inputs. The text description has no effect on the baseline. To activate the text guidance, `+ TAM' adds the text attention module (TAM) after all deconvolution layers of the baseline. In Table~\ref{tabnet}, the model `+ TAM' outputs images with better perceptual quality (NIQE/PI) and semantic accuracy (TIM). From Figure~\ref{fabnet}, we see that image details of `+ TAM' can be slightly changed by altering keywords in text descriptions. As expected, employing the TAM enables model to generate more textures related to text descriptions and to enhance the TIM score. Though SR results can correctly correspond to text descriptions, some real details are missing.

\begin{figure}[tb]
\centering \includegraphics[width=8.2cm]{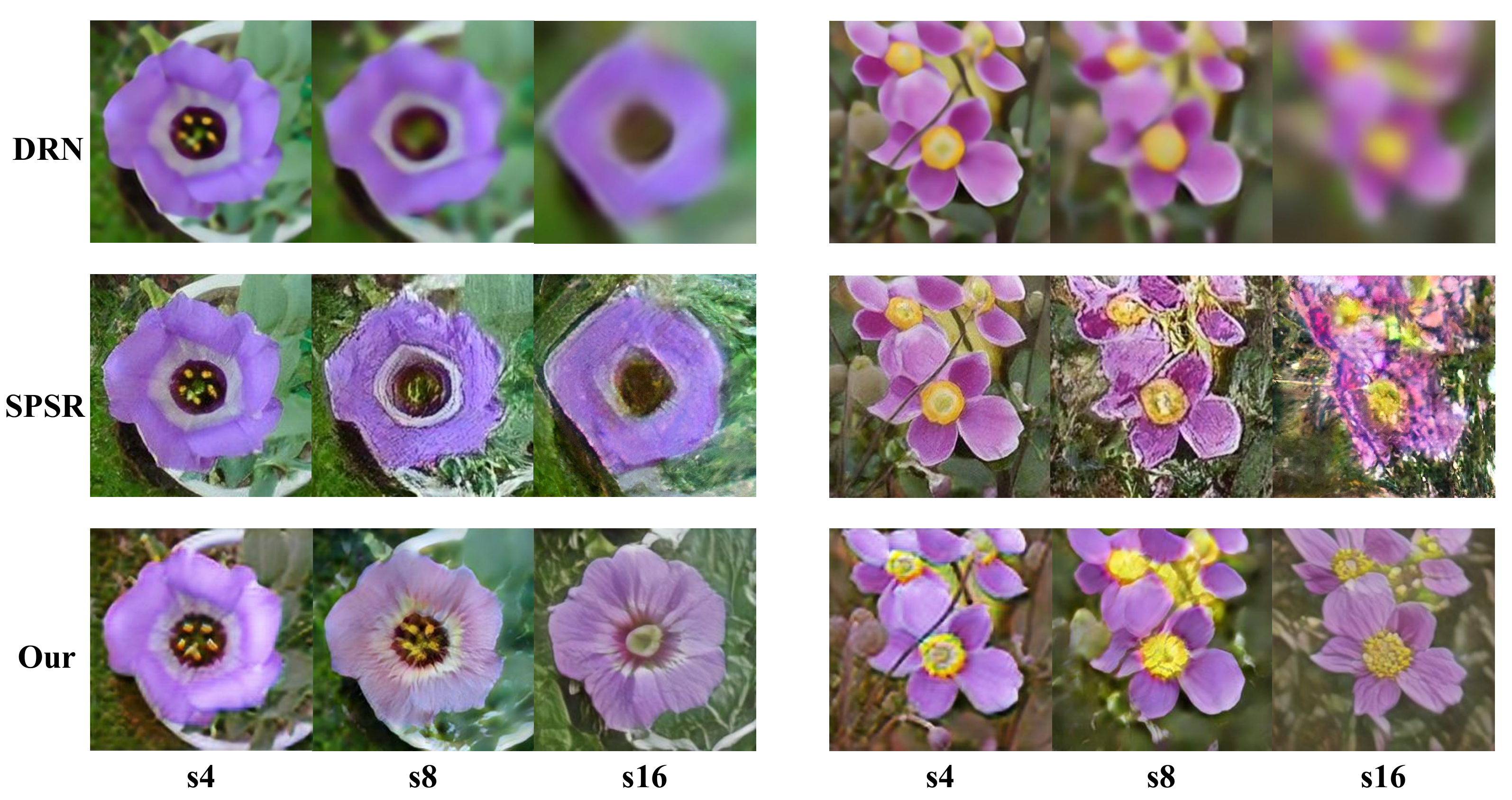} \vspace{-0.3em}
\caption{Visual comparison of different SR methods (EDSR~\cite{edsr}, SPSR~\cite{spsr}, and ours) on scale 4, 8, 16.} \label{fscale416} \end{figure}
\begin{figure}[tb]
\centering \includegraphics[width=0.43\textwidth]{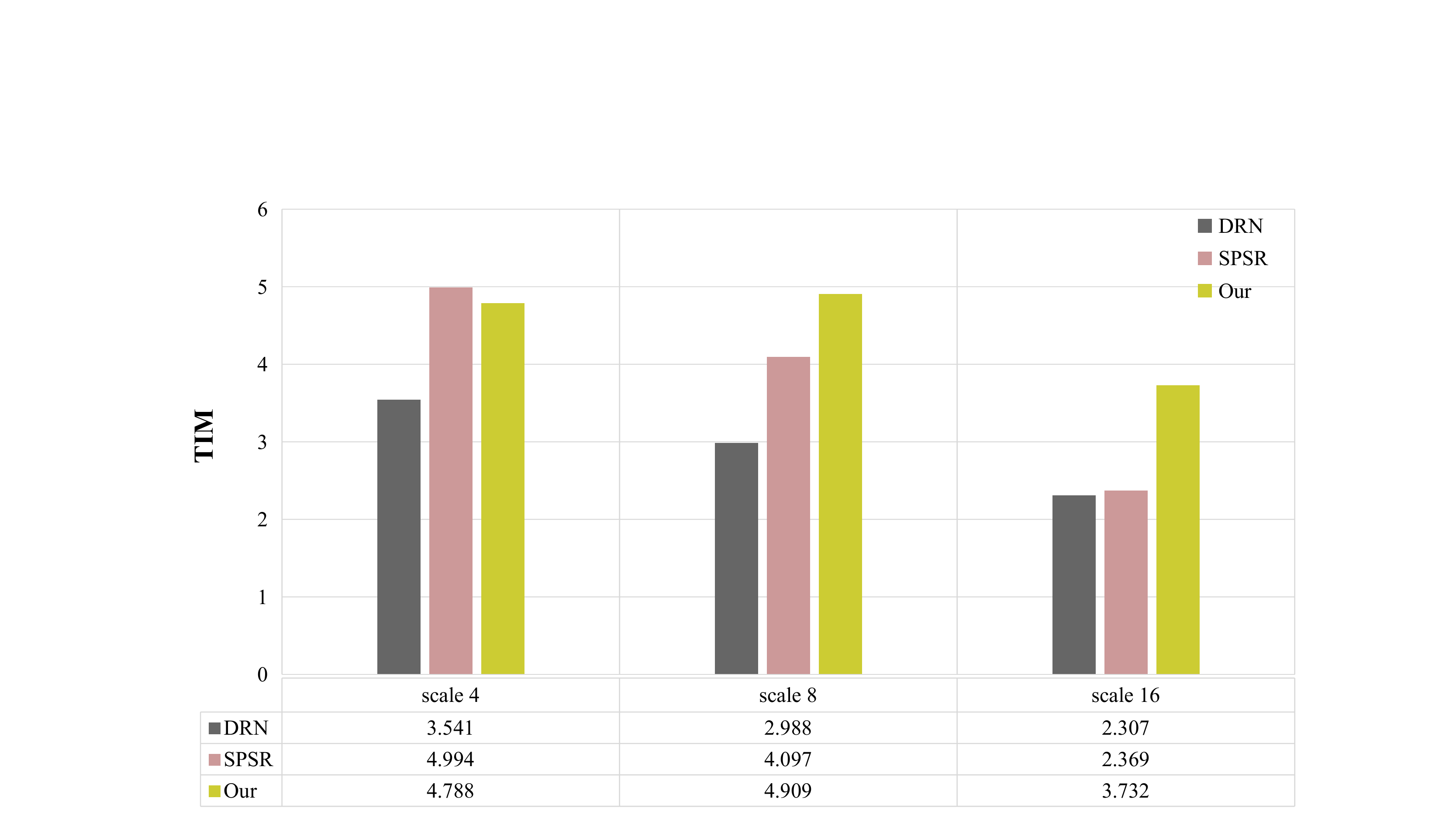} \vspace{-0.5em}
\caption{TIM variation of different SR methods on different scales. TIM scores reflect amounts of semantic information contained in SR images. The proposed TGSR keeps a well stability on different scales in terms of semantic accuracy.} \label{fscaletb} \end{figure}

After employing the text-image consistency loss ($\mathcal{L}_{TIC}$) on the `+ TAM' model, we obtain a model `+ $\mathcal{L}_{TIC}$'. In Table~\ref{tabnet}, the `+ $\mathcal{L}_{TIC}$' has similar NIQE/PI as the `+ TAM', but higher TIM, which illustrates that the loss $\mathcal{L}_{TIC}$ can improve the semantic accuracy of images relative to text. From the Figure~\ref{fabnet}, the `+ $\mathcal{L}_{TIC}$' restores more realistic bird textures.

To refine image details, the model `+ coarse-to-fine' introduces the coarse-to-fine structure by adding the refine branch based on the `+ $\mathcal{L}_{TIC}$' and regarding the `+ TAM' as the global branch. Lower NIQE/PI illustrates the refine branch can effectively promote visual SR results. The final model `+ $\mathcal{L}_{TAR}$', which further employs the text-attention reconstruction loss $\mathcal{L}_{TAR}$, obtains the best NIQE/PI and restores more authentic textures. As shown in Figure~\ref{fabnet}, the results of the `+ $\mathcal{L}_{TAR}$' have abundant and photo-realistic image details, such as the head and feather of the bird. 

\begin{table}[tb]
\centering \scalebox{0.76}[0.76]{ \begin{tabular}{c|cccc|c}
\toprule \toprule
metrics &Bicubic &SuperFAN~\cite{superfan}&DICGAN~\cite{dicgan}&Ours&GroundTruth \\
\midrule
TIM$\uparrow$ &0.049&0.013&0.378&\textbf{0.885}&0.815\\
PSNR$\uparrow$ &25.81&24.85&27.42&23.48&-\\
SSIM$\uparrow$ &0.844&0.859&0.862&0.766&-\\
NIQE$\downarrow$ &14.514&10.347&5.657&8.8464&7.473\\
PI$\downarrow$ &9.676&8.606&5.002&7.165&5.256\\
FID$\downarrow$ &105.232&120.666&92.901&93.919&- \\
\bottomrule \end{tabular} }
\caption{Quantitative comparison with state-of-the-art face SR methods on the CelebA dataset.}
\label{tstoaface}  \end{table}    

\begin{table}[tb]
  \centering  \scalebox{0.73}[0.73]{
    \begin{tabular}{c|c|ccccc|p{17mm}<{\centering}}
    \toprule   \toprule
     Dataset&metrics&Bicubic&EDSR& ESRGAN&SPSR&Ours&GroundTruth \\
 \midrule
    \multirow{3}[0]{*}{CUB}
    &TIM$\uparrow$& 0.920&1.090&2.482&2.045&\textbf{2.841}&3.189\\
    &NIQE$\downarrow$&12.374&10.684&\textbf{5.465}&5.885&6.623&6.734\\
    &PI$\downarrow$&9.747&8.168&2.644&3.345&\textbf{2.560}&2.302\\
 \midrule
    \multirow{3}[0]{*}{COCO}
    &TIM$\uparrow$&4.967&5.708&6.280&6.650&\textbf{7.353}&7.649\\
    &NIQE$\downarrow$&11.110&9.683&6.816&6.378&\textbf{6.4844}&3.840\\
    &PI$\downarrow$&9.373&8.515&7.135&6.060&\textbf{4.922}&3.657\\
 \bottomrule  \end{tabular}  } 
\caption{Comparison with state-of-the-arts on CUB and COCO datasets.} \label{tstoaother}  \end{table}

\begin{table}[!h]
  \centering  \scalebox{0.68}[0.68]{
    \begin{tabular}{cc|c|ccccc|p{17mm}<{\centering}}
    \toprule   \toprule
     Dataset&scale&metrics&Bicubic&EDSR&ESRGAN&SPSR&Ours&GroundTruth \\
 \midrule
    \multirow{3}{*}{Oxford-102}& \multirow{3}{*}{$\times$8}
    &TIM$\uparrow$&2.446&3.541&3.980&4.097&\textbf{4.778}&5.112\\
    &&NIQE$\downarrow$&11.089&9.405&4.860&\textbf{4.465}&5.282&4.506\\
&&PI$\downarrow$&10.333&8.564&4.183&\textbf{3.221}&5.100&4.688\\
   \midrule
    \multirow{3}{*}{Oxford-102}&\multirow{3}{*}{$\times$16}
    &TIM$\uparrow$&1.039&2.307&2.108&2.369&\textbf{3.732}&5.112\\
    &&NIQE$\downarrow$&13.495&11.685&7.541&5.993&\textbf{4.506}&4.159\\
   &&PI$\downarrow$&12.076&9.376&6.012&4.688&\textbf{3.654}&2.476\\
 \bottomrule   \end{tabular} } 
\caption{Quantitative comparison with state-of-the-arts on Oxford-102 dataset.} \label{tstoaother1}  \end{table}

\begin{figure}[!h]
\leftline{\small{The bird has \textbf{brown crown, striped feathers}.}}
\centering \subfigure{
\centering \begin{minipage}[t]{1.54cm} \centering \includegraphics[width=1.64cm]{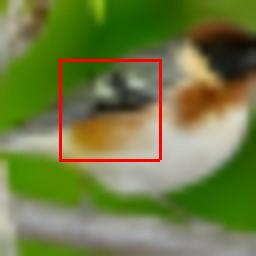} \end{minipage}
\centering \begin{minipage}[t]{1.54cm} \centering \includegraphics[width=1.64cm]{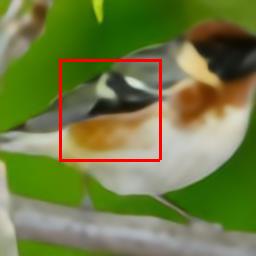}  \end{minipage}
\centering \begin{minipage}[t]{1.54cm} \centering \includegraphics[width=1.64cm]{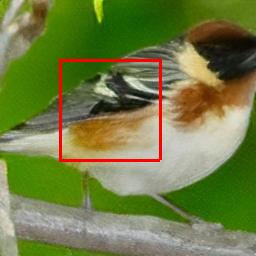} \end{minipage}
\centering \begin{minipage}[t]{1.54cm} \centering \includegraphics[width=1.64cm]{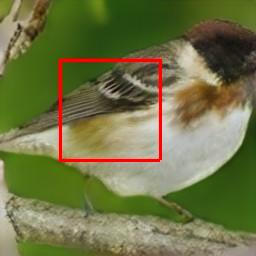} \end{minipage}
\centering \begin{minipage}[t]{1.54cm} \centering \includegraphics[width=1.64cm]{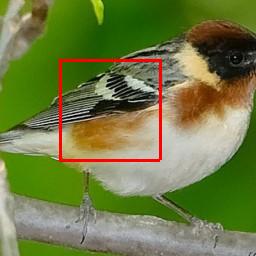} \end{minipage}
}\
\leftline{\small{A bird has \textbf{orange} belly and a \textbf{small black bill}.}}
\centering \subfigure{
 \begin{minipage}[t]{1.54cm} \centering  \includegraphics[width=1.64cm]{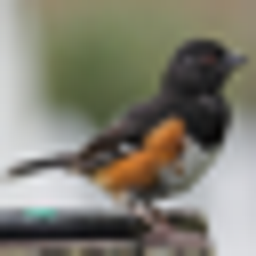} \end{minipage}
 \begin{minipage}[t]{1.54cm} \centering  \includegraphics[width=1.64cm]{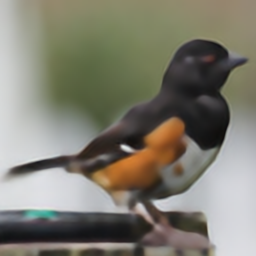} \end{minipage}
 \begin{minipage}[t]{1.54cm} \centering  \includegraphics[width=1.64cm]{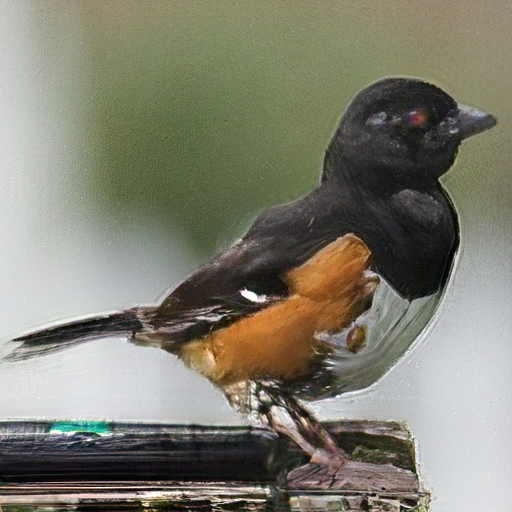} \end{minipage}
 \begin{minipage}[t]{1.54cm}  \centering \includegraphics[width=1.64cm]{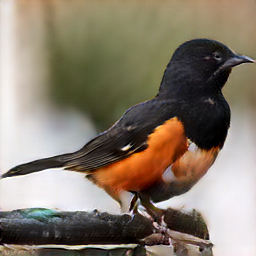} \end{minipage}
 \begin{minipage}[t]{1.54cm} \centering \includegraphics[width=1.64cm]{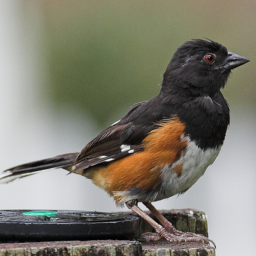} \end{minipage}
 }\
\leftline{\small{\textbf{Zebras} grazing in a field.}}
\centering \subfigure{
\begin{minipage}[t]{1.54cm} \centering  \includegraphics[width=1.64cm]{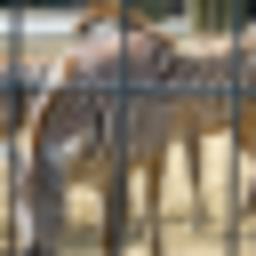}  \\ \centering{\small{Bicubic}}\end{minipage}
\begin{minipage}[t]{1.54cm} \centering  \includegraphics[width=1.64cm]{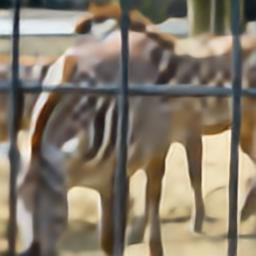} \\ \centering{\small{EDSR}} \end{minipage}
\begin{minipage}[t]{1.54cm} \centering  \includegraphics[width=1.64cm]{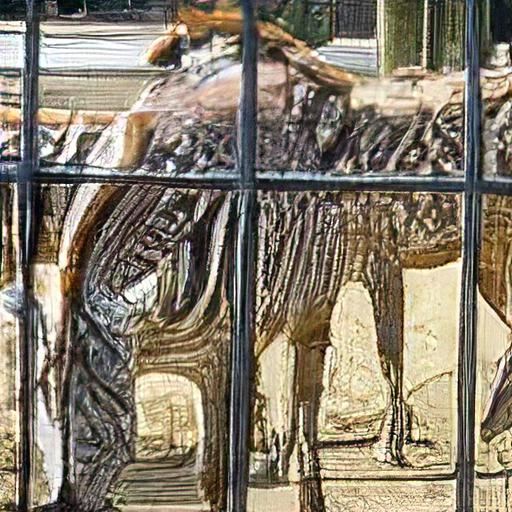}  \\ \centering{\small{SPSR}}\end{minipage}
\begin{minipage}[t]{1.54cm} \centering  \includegraphics[width=1.64cm]{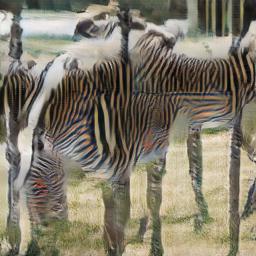} \\ \centering{\small{Ours}} \end{minipage}
\begin{minipage}[t]{1.54cm} \centering  \includegraphics[width=1.64cm]{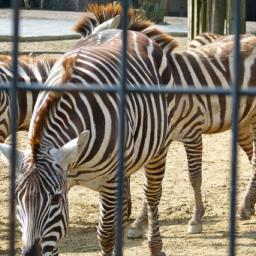} \\ \centering{\small{GT}} \end{minipage} }\  \vspace{-0.3em}
\caption{Comparison with SOTA on CUB and COCO.} \label{fsotacub} \end{figure}

\subsection{Comparison with State-of-the-arts} 
As we stated before, the network structure is not our main contribution, so we compare with several representative MSE-based and GAN-based SISR and Face SR methods, including EDSR~\cite{edsr}, ESRGAN~\cite{esrgan}, SPSR~\cite{spsr}, SuperFAN~\cite{superfan}, DICGAN~\cite{dicgan}, that have public codes and models. Note that most SR methods only provide $\times$4 models, we retrain $\times$8 SR models of the EDSR and ESRGAN on our training set for fair comparison by adding an additional $\times$2 upsampling layer at the end of their original models. 

\textbf{Stability to different scales:} The visual performance and the quantitative TIM variation of different SR methods at different scales are demonstrated in Figure~\ref{fscale416} and Figure~\ref{fscaletb}. We can observe that the performance drop is obvious for existing SR methods, facing larger scale factors.
In Figure~\ref{fscale416}, the compared SR methods (DRN and SPSR) either generate blur textures or fake artifacts for scale 8 and 16. The proposed TGSR can generate clear image details and accurate semantic information. In Figure~\ref{fscaletb}, the TIM results of DRN and SPSR decrease significantly accompanied by increasing scale factors. For a small scale, the LR image can provide enough information to restore semantically accurate images, therefore these SR models can well deal with factor by only using pixel-wise constraint and generate images, that are more consistent with the text descriptions and have higher TIM scores. For larger scale, these SR methods become less efficiency for LR image fail to provide enough information, which leads to worse visual quality and text-image matching score. In comparison, the proposed TGSR keeps stable TIM score, for it can restore the intelligibility of an image with accurate semantic information even for a large scale. The proposed TGSR has an obvious advantage for large scale factors.

Above experiments demonstrate that traditional single-modal SISR methods are significantly influenced by scale factors as they are trained to decrease the pixel-wise reconstruction error, and only utilize the spatial information from the LR image, and the further improvement of perceptual quality is restricted.

\textbf{Quantitative metrics:} Table~\ref{tstoaface}, \ref{tstoaother}, \ref{tstoaother1} report quantitative comparisons with state-of-the-art SISR methods and face SR methods. In these Tables, NIQE/PI of some GAN-based SR methods are lower than that of HR images, which goes against the common sense and demonstrates NIQE/PI cannot fairly evaluate GAN-based image restoration algorithms, as stated in~\cite{pipal}. The PSNR/SSIM are not appropriate to evaluate SR images in terms of the human perception. Our approach makes significant progress on TIM and obtains NIQE/PI closer to the GT.

\begin{figure}[tb]
\leftline{The \textbf{young} man is smiling.}
\centering \subfigure{
\begin{minipage}[t]{1.54cm} \centering  \includegraphics[width=1.64cm]{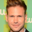} \end{minipage}
 \begin{minipage}[t]{1.54cm} \centering  \includegraphics[width=1.64cm]{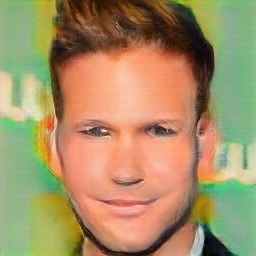}   \end{minipage}
\begin{minipage}[t]{1.54cm}  \centering \includegraphics[width=1.64cm]{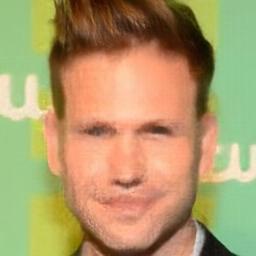} \end{minipage}
\begin{minipage}[t]{1.54cm} \centering  \includegraphics[width=1.64cm]{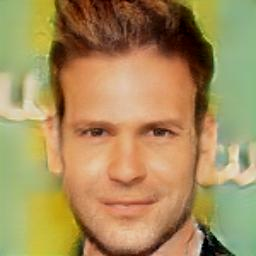} \end{minipage}
\begin{minipage}[t]{1.54cm}  \centering  \includegraphics[width=1.64cm]{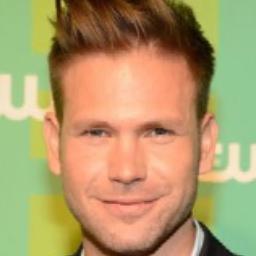} \end{minipage}
}\
\leftline{The \textbf{double chined} man has high cheekbones.}
\centering \subfigure{
\begin{minipage}[t]{1.54cm} \centering  \includegraphics[width=1.64cm]{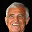} \end{minipage}
 \begin{minipage}[t]{1.54cm} \centering  \includegraphics[width=1.64cm]{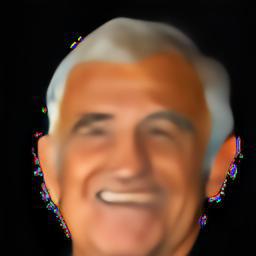} \end{minipage}
\begin{minipage}[t]{1.54cm}  \centering \includegraphics[width=1.64cm]{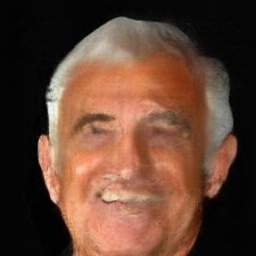} \end{minipage}
\begin{minipage}[t]{1.54cm} \centering  \includegraphics[width=1.64cm]{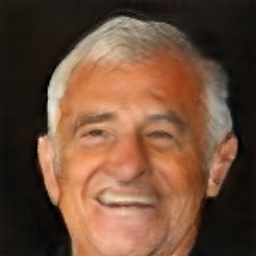} \end{minipage}
\begin{minipage}[t]{1.54cm}  \centering  \includegraphics[width=1.64cm]{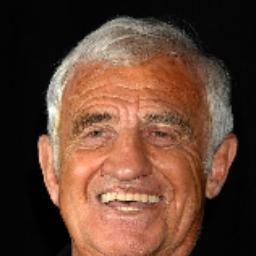} \end{minipage}
}\
\leftline{She has straight hair which is \textbf{blond}.}
\centering \subfigure{
\begin{minipage}[t]{1.54cm} \centering  \includegraphics[width=1.64cm]{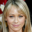}  \\ \centering{\small{LR}}
 \end{minipage}
\begin{minipage}[t]{1.54cm} \centering  \includegraphics[width=1.64cm]{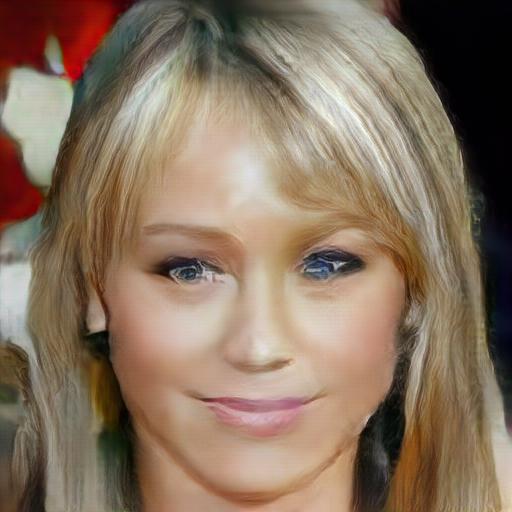}\\ \centering{\small{SuperFAN}}  \end{minipage}
\begin{minipage}[t]{1.54cm}  \centering \includegraphics[width=1.64cm]{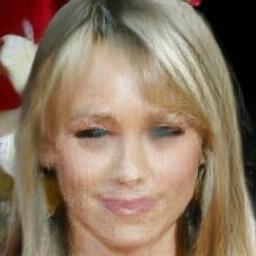}  \\ \centering{\small{DICGAN}}\end{minipage}
\begin{minipage}[t]{1.54cm} \centering  \includegraphics[width=1.64cm]{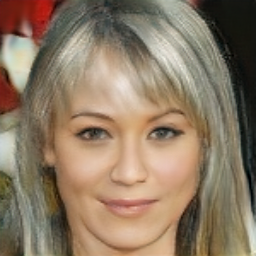} \\ \centering{Ours} \end{minipage}
\begin{minipage}[t]{1.54cm}  \centering \includegraphics[width=1.64cm]{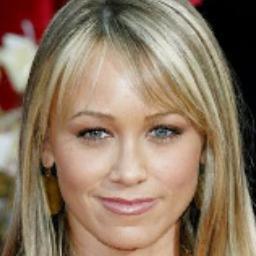}  \\ \centering{\small{GT}} \end{minipage} }\ \vspace{-0.6em}
\caption{Visual comparison with state-of-the-art face SR methods on the CelebA. The text inputs are above images.}
\label{fsotaface}  \end{figure}

\begin{figure}[tb]
\centering \subfigure{
\begin{minipage}[t]{1.8cm} \centering  \includegraphics[width=1.88cm]{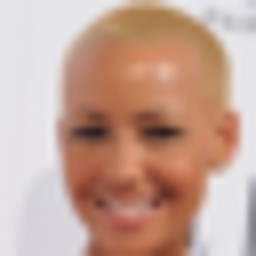}  \\ \centering{\small{Bicubic}} \end{minipage}
\begin{minipage}[t]{1.8cm} \centering \includegraphics[width=1.88cm]{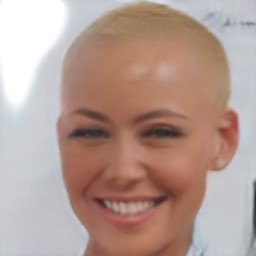} \\ \centering{\tiny{\textbf{arched eyebrows}}} \end{minipage}
\begin{minipage}[t]{1.8cm} \centering \includegraphics[width=1.88cm]{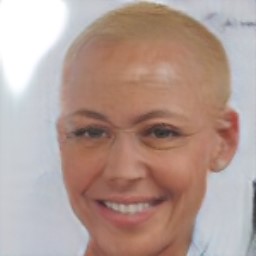} \\ \centering{\tiny{\textbf{old man}}} \end{minipage}
\begin{minipage}[t]{1.8cm} \centering \includegraphics[width=1.88cm]{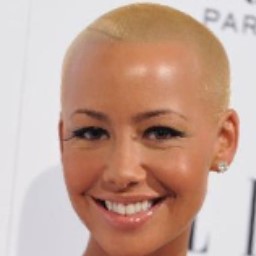} \\ \centering{\tiny{GT}} \end{minipage}
}\ \centering \subfigure{
\begin{minipage}[t]{1.8cm} \centering \includegraphics[width=1.88cm]{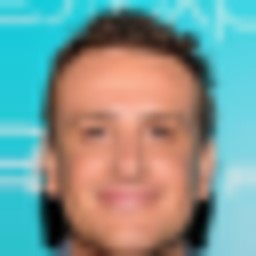}  \\ \centering{\small{Bicubic}} \end{minipage}
\begin{minipage}[t]{1.8cm}  \centering \includegraphics[width=1.88cm]{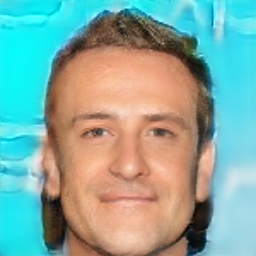}  \\ \centering{\tiny{\textbf{high cheekbones}}} \end{minipage}
\begin{minipage}[t]{1.8cm} \centering \includegraphics[width=1.88cm]{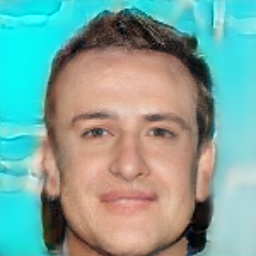} \\ \centering{\tiny{\textbf{attractive, pale skin}}} \end{minipage}
\begin{minipage}[t]{1.8cm} \centering \includegraphics[width=1.88cm]{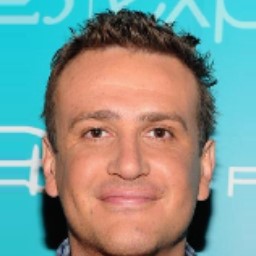} \\ \centering{\tiny{GT}} \end{minipage}
}\
\centering  \subfigure{
\begin{minipage}[t]{1.8cm} \centering  \includegraphics[width=1.88cm]{f11bic.jpg}  \\ \centering{\small{Bicubic}} \end{minipage}
\begin{minipage}[t]{1.8cm} \centering \includegraphics[width=1.88cm]{f11our1.jpg}  \\ \centering{\tiny{\textbf{Zebras}}} \end{minipage}
\begin{minipage}[t]{1.8cm} \centering \includegraphics[width=1.88cm]{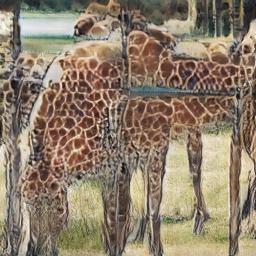}\\ \centering{\tiny{\textbf{Giraffes}}} \end{minipage}
\begin{minipage}[t]{1.8cm} \centering \includegraphics[width=1.88cm]{f11hr.jpg} \\ \centering{\tiny{GT}} \end{minipage}
}\ \centering \subfigure{
\begin{minipage}[t]{1.8cm} \centering \includegraphics[width=1.88cm]{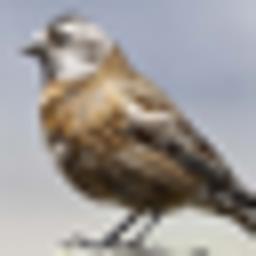}\\ \centering{\small{Bicubic}}
\end{minipage}
\begin{minipage}[t]{1.8cm} \centering \includegraphics[width=1.88cm]{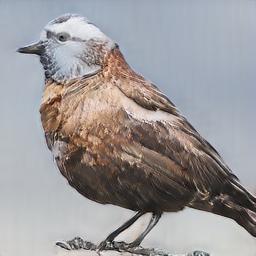}\\ \centering{\tiny{\textbf{white head}.}}  \end{minipage}
\begin{minipage}[t]{1.8cm} \centering \includegraphics[width=1.88cm]{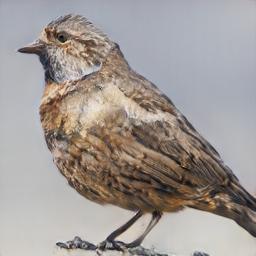} \\ \centering{\tiny{\textbf{brown head}.}} \end{minipage}
\begin{minipage}[t]{1.8cm} \centering \includegraphics[width=1.88cm]{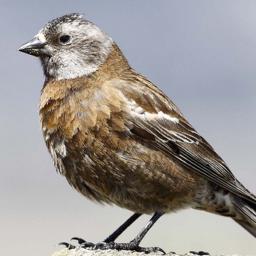} \\ \centering{\tiny{GT}} \end{minipage}
}\ \vspace{-0.6em}
\caption{Examples of manipulating details in SR results with different keywords in text descriptions.}  \label{ftxt} \end{figure}

\textbf{Visual results:} Figure~\ref{f0}, \ref{fsotacub}, and \ref{fsotaface} demonstrate the subjective SR results. It is difficult for traditional SR networks to predict accurate HR contents by only referring to the finite pixel information in tiny LR images. In comparison, the proposed TGSR exploits the text guidance to restore better visual results and faithful textures (e.g., the feathers of birds) that are consistent with text descriptions.

\subsection{Editable Super-Resolution} 
We present one interesting application of TGSR, a text-guided editable image reconstruction, where users can manually edit SR results based on the text input. The possibility to tweak a model's output is important for making super-resolution more useful and controllable. As shown in Figure~\ref{ftxt}, the proposed text guided SR model is able to flexibly control image details, such as skin, texture, age, and generates different HR image contents based on different keywords in text descriptions.

\textbf{Limitation:} The main contribution of this paper is introducing the text descriptions in SISR. We observe some limitations in our work. First, since text descriptions in existing datasets may not cover enough details, the text becomes invalid sometimes. Besides, text may correspond to multi visual representations, which influences fidelity of the final result. 

\section{Conclusion}
This paper aims to solve the larger factor super-resolution by regarding the single image super-resolution problem as a text-guided detail generation problem, and proposes a multimodal text-guided super-resolution algorithm, which restores realistic and rational image details by utilizing the guidance of text descriptions in a coarse-to-fine process. The text-guided losses enable the network to focus on objects concentrated by the text descriptions, and keep the consistency between the super-resolved image and the corresponding text description. Experimental results demonstrate that the proposed approach can flexibly incorporate the text prior to facilitate the rich-detail super-resolution, which is practical in reality.

\bibliographystyle{ACM-Reference-Format}
\bibliography{sample-base}


\begin{thebibliography}{37}


\ifx \showCODEN    \undefined \def \showCODEN     #1{\unskip}     \fi
\ifx \showDOI      \undefined \def \showDOI       #1{#1}\fi
\ifx \showISBNx    \undefined \def \showISBNx     #1{\unskip}     \fi
\ifx \showISBNxiii \undefined \def \showISBNxiii  #1{\unskip}     \fi
\ifx \showISSN     \undefined \def \showISSN      #1{\unskip}     \fi
\ifx \showLCCN     \undefined \def \showLCCN      #1{\unskip}     \fi
\ifx \shownote     \undefined \def \shownote      #1{#1}          \fi
\ifx \showarticletitle \undefined \def \showarticletitle #1{#1}   \fi
\ifx \showURL      \undefined \def \showURL       {\relax}        \fi
\providecommand\bibfield[2]{#2}
\providecommand\bibinfo[2]{#2}
\providecommand\natexlab[1]{#1}
\providecommand\showeprint[2][]{arXiv:#2}

\bibitem[\protect\citeauthoryear{Bahat and Michaeli}{Bahat and
  Michaeli}{2020}]%
        {explorable}
\bibfield{author}{\bibinfo{person}{Yuval Bahat} {and} \bibinfo{person}{Tomer
  Michaeli}.} \bibinfo{year}{2020}\natexlab{}.
\newblock \showarticletitle{Explorable Super Resolution}. In
  \bibinfo{booktitle}{\emph{IEEE/CVF Conference on Computer Vision and Pattern
  Recognition (CVPR)}}.
\newblock


\bibitem[\protect\citeauthoryear{Bulat and Tzimiropoulos}{Bulat and
  Tzimiropoulos}{2018}]%
        {superfan}
\bibfield{author}{\bibinfo{person}{Adrian Bulat} {and}
  \bibinfo{person}{Georgios Tzimiropoulos}.} \bibinfo{year}{2018}\natexlab{}.
\newblock \showarticletitle{Super-FAN: Integrated facial landmark localization
  and super-resolution of real-world low resolution faces in arbitrary poses
  with GANs}. In \bibinfo{booktitle}{\emph{IEEE/CVF Conference on Computer
  Vision and Pattern Recognition (CVPR)}}.
\newblock


\bibitem[\protect\citeauthoryear{Bühler, Romero, and Timofte}{Bühler
  et~al\mbox{.}}{2020}]%
        {deepsee}
\bibfield{author}{\bibinfo{person}{Marcel~Christoph Bühler},
  \bibinfo{person}{Andrés Romero}, {and} \bibinfo{person}{Radu Timofte}.}
  \bibinfo{year}{2020}\natexlab{}.
\newblock \showarticletitle{DeepSEE: Deep Disentangled Semantic Explorative
  Extreme Super-Resolution}.
\newblock


\bibitem[\protect\citeauthoryear{Chen, Shen, Gao, Liu, and Liu}{Chen
  et~al\mbox{.}}{2018}]%
        {lbieram}
\bibfield{author}{\bibinfo{person}{Jianbo Chen}, \bibinfo{person}{Yelong Shen},
  \bibinfo{person}{Jianfeng Gao}, \bibinfo{person}{Jingjing Liu}, {and}
  \bibinfo{person}{Xiaodong Liu}.} \bibinfo{year}{2018}\natexlab{}.
\newblock \showarticletitle{Language-Based Image Editing with Recurrent
  Attentive Models}. In \bibinfo{booktitle}{\emph{IEEE/CVF Conference on
  Computer Vision and Pattern Recognition (CVPR)}}.
\newblock


\bibitem[\protect\citeauthoryear{{Chen}, {Tai}, {Liu}, {Shen}, and
  {Yang}}{{Chen} et~al\mbox{.}}{2018}]%
        {fsrnet}
\bibfield{author}{\bibinfo{person}{Y. {Chen}}, \bibinfo{person}{Y. {Tai}},
  \bibinfo{person}{X. {Liu}}, \bibinfo{person}{C. {Shen}}, {and}
  \bibinfo{person}{J. {Yang}}.} \bibinfo{year}{2018}\natexlab{}.
\newblock \showarticletitle{FSRNet: End-to-End Learning Face Super-Resolution
  with Facial Priors}. In \bibinfo{booktitle}{\emph{IEEE/CVF Conference on
  Computer Vision and Pattern Recognition (CVPR)}}.
  \bibinfo{pages}{2492--2501}.
\newblock


\bibitem[\protect\citeauthoryear{Cho, Bahng, Park, Yoo, Wu, Ma, and Choo}{Cho
  et~al\mbox{.}}{2018}]%
        {Text2Colors}
\bibfield{author}{\bibinfo{person}{Wonwoong Cho}, \bibinfo{person}{Hyojin
  Bahng}, \bibinfo{person}{David~Keetae Park}, \bibinfo{person}{Seungjoo Yoo},
  \bibinfo{person}{Ziming Wu}, \bibinfo{person}{Xiaojuan Ma}, {and}
  \bibinfo{person}{Jaegul Choo}.} \bibinfo{year}{2018}\natexlab{}.
\newblock \showarticletitle{Text2Colors: Guiding Image Colorization through
  Text-Driven Palette Generation}. In \bibinfo{booktitle}{\emph{European
  Conference on Computer Vision (ECCV)}}.
\newblock


\bibitem[\protect\citeauthoryear{Dong, Loy, He, and Tang}{Dong
  et~al\mbox{.}}{2016}]%
        {srcnn}
\bibfield{author}{\bibinfo{person}{C. Dong}, \bibinfo{person}{C.~C. Loy},
  \bibinfo{person}{K. He}, {and} \bibinfo{person}{X. Tang}.}
  \bibinfo{year}{2016}\natexlab{}.
\newblock \showarticletitle{Image Super-Resolution Using Deep Convolutional
  Networks}.
\newblock \bibinfo{journal}{\emph{IEEE Transactions on Pattern Analysis and
  Machine Intelligence}} \bibinfo{volume}{38}, \bibinfo{number}{2}
  (\bibinfo{date}{Feb} \bibinfo{year}{2016}), \bibinfo{pages}{295--307}.
\newblock


\bibitem[\protect\citeauthoryear{Goodfellow, Pouget-Abadie, Mirza, Xu,
  Warde-Farley, Ozair, Courville, and Bengio}{Goodfellow et~al\mbox{.}}{2014}]%
        {gan}
\bibfield{author}{\bibinfo{person}{Ian~J. Goodfellow}, \bibinfo{person}{Jean
  Pouget-Abadie}, \bibinfo{person}{Mehdi Mirza}, \bibinfo{person}{Bing Xu},
  \bibinfo{person}{David Warde-Farley}, \bibinfo{person}{Sherjil Ozair},
  \bibinfo{person}{Aaron Courville}, {and} \bibinfo{person}{Yoshua Bengio}.}
  \bibinfo{year}{2014}\natexlab{}.
\newblock \showarticletitle{Generative Adversarial Networks}.
\newblock \bibinfo{journal}{\emph{Advances in Neural Information Processing
  Systems}}  \bibinfo{volume}{3} (\bibinfo{year}{2014}),
  \bibinfo{pages}{2672--2680}.
\newblock


\bibitem[\protect\citeauthoryear{Gu, Cai, Chen, Ye, Ren, and Dong}{Gu
  et~al\mbox{.}}{2020}]%
        {pipal}
\bibfield{author}{\bibinfo{person}{Jinjin Gu}, \bibinfo{person}{Haoming Cai},
  \bibinfo{person}{Haoyu Chen}, \bibinfo{person}{Xiaoxing Ye},
  \bibinfo{person}{Jimmy Ren}, {and} \bibinfo{person}{Chao Dong}.}
  \bibinfo{year}{2020}\natexlab{}.
\newblock \showarticletitle{PIPAL: a Large-Scale Image Quality Assessment
  Dataset for Perceptual Image Restoration}. In
  \bibinfo{booktitle}{\emph{European Conference on Computer Vision (ECCV)}}.
\newblock


\bibitem[\protect\citeauthoryear{Guo, Zhu, Zhao, Cao, Lei, and Li}{Guo
  et~al\mbox{.}}{2020b}]%
        {audioSR}
\bibfield{author}{\bibinfo{person}{Jianzhu Guo}, \bibinfo{person}{Xiangyu Zhu},
  \bibinfo{person}{Chenxu Zhao}, \bibinfo{person}{Dong Cao},
  \bibinfo{person}{Zhen Lei}, {and} \bibinfo{person}{Stan~Z Li}.}
  \bibinfo{year}{2020}\natexlab{b}.
\newblock \showarticletitle{Learning Meta Face Recognition in Unseen Domains}.
  In \bibinfo{booktitle}{\emph{IEEE/CVF Conference on Computer Vision and
  Pattern Recognition (CVPR)}}.
\newblock


\bibitem[\protect\citeauthoryear{Guo, Chen, Wang, Chen, Cao, Deng, Xu, and
  Tan}{Guo et~al\mbox{.}}{2020a}]%
        {drn}
\bibfield{author}{\bibinfo{person}{Yong Guo}, \bibinfo{person}{Jian Chen},
  \bibinfo{person}{Jingdong Wang}, \bibinfo{person}{Qi Chen},
  \bibinfo{person}{Jiezhang Cao}, \bibinfo{person}{Zeshuai Deng},
  \bibinfo{person}{Yanwu Xu}, {and} \bibinfo{person}{Mingkui Tan}.}
  \bibinfo{year}{2020}\natexlab{a}.
\newblock \showarticletitle{Closed-loop Matters: Dual Regression Networks for
  Single Image Super-Resolution}. In \bibinfo{booktitle}{\emph{IEEE/CVF
  Conference on Computer Vision and Pattern Recognition (CVPR)}}.
\newblock


\bibitem[\protect\citeauthoryear{Kim, Jhoo, Park, and Yoo}{Kim
  et~al\mbox{.}}{2019}]%
        {Tag2Pix}
\bibfield{author}{\bibinfo{person}{Hyunsu Kim}, \bibinfo{person}{Ho~Young
  Jhoo}, \bibinfo{person}{Eunhyeok Park}, {and} \bibinfo{person}{Sungjoo Yoo}.}
  \bibinfo{year}{2019}\natexlab{}.
\newblock \showarticletitle{Tag2Pix: Line Art Colorization Using Text Tag With
  SECat and Changing Loss}. In \bibinfo{booktitle}{\emph{IEEE International
  Conference on Computer Vision (ICCV)}}.
\newblock


\bibitem[\protect\citeauthoryear{Kim, Lee, and Lee}{Kim et~al\mbox{.}}{2016}]%
        {vdsr}
\bibfield{author}{\bibinfo{person}{J. Kim}, \bibinfo{person}{J.~K. Lee}, {and}
  \bibinfo{person}{K.~M. Lee}.} \bibinfo{year}{2016}\natexlab{}.
\newblock \showarticletitle{Accurate Image Super-Resolution Using Very Deep
  Convolutional Networks}. In \bibinfo{booktitle}{\emph{IEEE/CVF Conference on
  Computer Vision and Pattern Recognition (CVPR)}}.
  \bibinfo{pages}{1646--1654}.
\newblock


\bibitem[\protect\citeauthoryear{Ledig, Theis, Huszár, Caballero, Cunningham,
  Acosta, Aitken, Tejani, Totz, Wang, and Shi}{Ledig et~al\mbox{.}}{2017}]%
        {srgan}
\bibfield{author}{\bibinfo{person}{Christian Ledig}, \bibinfo{person}{Lucas
  Theis}, \bibinfo{person}{Ferenc Huszár}, \bibinfo{person}{Jose Caballero},
  \bibinfo{person}{Andrew Cunningham}, \bibinfo{person}{Alejandro Acosta},
  \bibinfo{person}{Andrew Aitken}, \bibinfo{person}{Alykhan Tejani},
  \bibinfo{person}{Johannes Totz}, \bibinfo{person}{Zehan Wang}, {and}
  \bibinfo{person}{Wenzhe Shi}.} \bibinfo{year}{2017}\natexlab{}.
\newblock \showarticletitle{Photo-Realistic Single Image Super-Resolution Using
  a Generative Adversarial Network}. In \bibinfo{booktitle}{\emph{IEEE/CVF
  Conference on Computer Vision and Pattern Recognition (CVPR)}}.
  \bibinfo{pages}{4681--4690}.
\newblock


\bibitem[\protect\citeauthoryear{Li, Qi, Lukasiewicz, and Torr}{Li
  et~al\mbox{.}}{2020}]%
        {manigan}
\bibfield{author}{\bibinfo{person}{Bowen Li}, \bibinfo{person}{Xiaojuan Qi},
  \bibinfo{person}{Thomas Lukasiewicz}, {and} \bibinfo{person}{Philip~HS
  Torr}.} \bibinfo{year}{2020}\natexlab{}.
\newblock \showarticletitle{Manigan: Text-guided image manipulation}. In
  \bibinfo{booktitle}{\emph{IEEE/CVF Conference on Computer Vision and Pattern
  Recognition (CVPR)}}. \bibinfo{pages}{7880--7889}.
\newblock


\bibitem[\protect\citeauthoryear{Li, Zhang, Zhang, Huang, He, Lyu, and Gao}{Li
  et~al\mbox{.}}{2019}]%
        {textimgan}
\bibfield{author}{\bibinfo{person}{Wenbo Li}, \bibinfo{person}{Pengchuan
  Zhang}, \bibinfo{person}{Lei Zhang}, \bibinfo{person}{Qiuyuan Huang},
  \bibinfo{person}{Xiaodong He}, \bibinfo{person}{Siwei Lyu}, {and}
  \bibinfo{person}{Jianfeng Gao}.} \bibinfo{year}{2019}\natexlab{}.
\newblock \showarticletitle{Object-driven Text-to-Image Synthesis via
  Adversarial Training}. In \bibinfo{booktitle}{\emph{IEEE/CVF Conference on
  Computer Vision and Pattern Recognition (CVPR)}}.
\newblock


\bibitem[\protect\citeauthoryear{Lim, Son, Kim, Nah, and Lee}{Lim
  et~al\mbox{.}}{2017}]%
        {edsr}
\bibfield{author}{\bibinfo{person}{Bee Lim}, \bibinfo{person}{Sanghyun Son},
  \bibinfo{person}{Heewon Kim}, \bibinfo{person}{Seungjun Nah}, {and}
  \bibinfo{person}{Kyoung~Mu Lee}.} \bibinfo{year}{2017}\natexlab{}.
\newblock \showarticletitle{Enhanced deep residual networks for single image
  super-resolution}. In \bibinfo{booktitle}{\emph{IEEE/CVF Conference on
  Computer Vision and Pattern Recognition (CVPR)}}, Vol.~\bibinfo{volume}{1}.
  \bibinfo{pages}{4}.
\newblock


\bibitem[\protect\citeauthoryear{Lin, Maire, Belongie, Hays, and Zitnick}{Lin
  et~al\mbox{.}}{2014}]%
        {coco}
\bibfield{author}{\bibinfo{person}{Tsung~Yi Lin}, \bibinfo{person}{Michael
  Maire}, \bibinfo{person}{Serge Belongie}, \bibinfo{person}{James Hays}, {and}
  \bibinfo{person}{C.~Lawrence Zitnick}.} \bibinfo{year}{2014}\natexlab{}.
\newblock \showarticletitle{Microsoft COCO: Common Objects in Context}. In
  \bibinfo{booktitle}{\emph{European Conference on Computer Vision (ECCV)}}.
\newblock


\bibitem[\protect\citeauthoryear{Liu, Zhang, Tang, Tang, and Wu}{Liu
  et~al\mbox{.}}{2020}]%
        {rfa}
\bibfield{author}{\bibinfo{person}{Jie Liu}, \bibinfo{person}{Wenjie Zhang},
  \bibinfo{person}{Yuting Tang}, \bibinfo{person}{Jie Tang}, {and}
  \bibinfo{person}{Gangshan Wu}.} \bibinfo{year}{2020}\natexlab{}.
\newblock \showarticletitle{Residual Feature Aggregation Network for Image
  Super-Resolution}. In \bibinfo{booktitle}{\emph{IEEE/CVF Conference on
  Computer Vision and Pattern Recognition (CVPR)}}.
\newblock


\bibitem[\protect\citeauthoryear{{Liu}, {Luo}, {Wang}, and {Tang}}{{Liu}
  et~al\mbox{.}}{2015}]%
        {celeba}
\bibfield{author}{\bibinfo{person}{Z. {Liu}}, \bibinfo{person}{P. {Luo}},
  \bibinfo{person}{X. {Wang}}, {and} \bibinfo{person}{X. {Tang}}.}
  \bibinfo{year}{2015}\natexlab{}.
\newblock \showarticletitle{Deep Learning Face Attributes in the Wild}. In
  \bibinfo{booktitle}{\emph{IEEE International Conference on Computer Vision
  (ICCV)}}. \bibinfo{pages}{3730--3738}.
\newblock
\urldef\tempurl%
\url{https://doi.org/10.1109/ICCV.2015.425}
\showDOI{\tempurl}


\bibitem[\protect\citeauthoryear{Ma, Jiang, Rao, Lu, and Zhou}{Ma
  et~al\mbox{.}}{2020a}]%
        {dicgan}
\bibfield{author}{\bibinfo{person}{Cheng Ma}, \bibinfo{person}{Zhenyu Jiang},
  \bibinfo{person}{Yongming Rao}, \bibinfo{person}{Jiwen Lu}, {and}
  \bibinfo{person}{Jie Zhou}.} \bibinfo{year}{2020}\natexlab{a}.
\newblock \showarticletitle{Deep Face Super-Resolution with Iterative
  Collaboration between Attentive Recovery and Landmark Estimation}. In
  \bibinfo{booktitle}{\emph{IEEE/CVF Conference on Computer Vision and Pattern
  Recognition (CVPR)}}.
\newblock


\bibitem[\protect\citeauthoryear{Ma, Rao, Cheng, Chen, Lu, and Zhou}{Ma
  et~al\mbox{.}}{2020b}]%
        {spsr}
\bibfield{author}{\bibinfo{person}{Cheng Ma}, \bibinfo{person}{Yongming Rao},
  \bibinfo{person}{Yean Cheng}, \bibinfo{person}{Ce Chen},
  \bibinfo{person}{Jiwen Lu}, {and} \bibinfo{person}{Jie Zhou}.}
  \bibinfo{year}{2020}\natexlab{b}.
\newblock \showarticletitle{Structure-Preserving Super Resolution with Gradient
  Guidance}. In \bibinfo{booktitle}{\emph{IEEE/CVF Conference on Computer
  Vision and Pattern Recognition (CVPR)}}.
\newblock


\bibitem[\protect\citeauthoryear{Menon, Damian, Hu, Ravi, and Rudin}{Menon
  et~al\mbox{.}}{2020}]%
        {pulse}
\bibfield{author}{\bibinfo{person}{Sachit Menon}, \bibinfo{person}{Alexandru
  Damian}, \bibinfo{person}{Shijia Hu}, \bibinfo{person}{Nikhil Ravi}, {and}
  \bibinfo{person}{Cynthia Rudin}.} \bibinfo{year}{2020}\natexlab{}.
\newblock \showarticletitle{{PULSE:} Self-Supervised Photo Upsampling via
  Latent Space Exploration of Generative Models}. In
  \bibinfo{booktitle}{\emph{IEEE/CVF Conference on Computer Vision and Pattern
  Recognition (CVPR)}}.
\newblock


\bibitem[\protect\citeauthoryear{{Mittal}, {Soundararajan}, and
  {Bovik}}{{Mittal} et~al\mbox{.}}{2013}]%
        {niqe}
\bibfield{author}{\bibinfo{person}{A. {Mittal}}, \bibinfo{person}{R.
  {Soundararajan}}, {and} \bibinfo{person}{A.~C. {Bovik}}.}
  \bibinfo{year}{2013}\natexlab{}.
\newblock \showarticletitle{Making a “Completely Blind” Image Quality
  Analyzer}.
\newblock \bibinfo{journal}{\emph{IEEE Signal Processing Letters}}
  \bibinfo{volume}{20}, \bibinfo{number}{3} (\bibinfo{year}{2013}),
  \bibinfo{pages}{209--212}.
\newblock


\bibitem[\protect\citeauthoryear{Nilsback and Zisserman}{Nilsback and
  Zisserman}{2008}]%
        {flower}
\bibfield{author}{\bibinfo{person}{M-E. Nilsback} {and} \bibinfo{person}{A.
  Zisserman}.} \bibinfo{year}{2008}\natexlab{}.
\newblock \showarticletitle{Automated Flower Classification over a Large Number
  of Classes}. In \bibinfo{booktitle}{\emph{Proceedings of the Indian
  Conference on Computer Vision, Graphics and Image Processing}}.
\newblock


\bibitem[\protect\citeauthoryear{Reed, Akata, Yan, Logeswaran, Schiele, and
  Lee}{Reed et~al\mbox{.}}{2016}]%
        {reed}
\bibfield{author}{\bibinfo{person}{Scott Reed}, \bibinfo{person}{Zeynep Akata},
  \bibinfo{person}{Xinchen Yan}, \bibinfo{person}{Lajanugen Logeswaran},
  \bibinfo{person}{Bernt Schiele}, {and} \bibinfo{person}{Honglak Lee}.}
  \bibinfo{year}{2016}\natexlab{}.
\newblock \showarticletitle{Generative Adversarial Text-to-Image Synthesis}. In
  \bibinfo{booktitle}{\emph{International Conference on Machine Learning
  (ICML)}}.
\newblock


\bibitem[\protect\citeauthoryear{{Schuster} and {Paliwal}}{{Schuster} and
  {Paliwal}}{1997}]%
        {lstm}
\bibfield{author}{\bibinfo{person}{M. {Schuster}} {and} \bibinfo{person}{K.~K.
  {Paliwal}}.} \bibinfo{year}{1997}\natexlab{}.
\newblock \showarticletitle{Bidirectional recurrent neural networks}.
\newblock \bibinfo{journal}{\emph{IEEE Transactions on Signal Processing}}
  \bibinfo{volume}{45}, \bibinfo{number}{11} (\bibinfo{year}{1997}),
  \bibinfo{pages}{2673--2681}.
\newblock
\urldef\tempurl%
\url{https://doi.org/10.1109/78.650093}
\showDOI{\tempurl}


\bibitem[\protect\citeauthoryear{Shen, Lai, Xu, Kautz, and Yang}{Shen
  et~al\mbox{.}}{2018}]%
        {shen}
\bibfield{author}{\bibinfo{person}{Ziyi Shen}, \bibinfo{person}{Wei{-}Sheng
  Lai}, \bibinfo{person}{Tingfa Xu}, \bibinfo{person}{Jan Kautz}, {and}
  \bibinfo{person}{Ming{-}Hsuan Yang}.} \bibinfo{year}{2018}\natexlab{}.
\newblock \showarticletitle{Deep Semantic Face Deblurring}.
\newblock \bibinfo{journal}{\emph{CoRR}}  \bibinfo{volume}{abs/1803.03345}
  (\bibinfo{year}{2018}).
\newblock


\bibitem[\protect\citeauthoryear{Szegedy, Vanhoucke, Ioffe, Shlens, and
  Wojna}{Szegedy et~al\mbox{.}}{2016}]%
        {inceptionv3}
\bibfield{author}{\bibinfo{person}{Christian Szegedy}, \bibinfo{person}{Vincent
  Vanhoucke}, \bibinfo{person}{Sergey Ioffe}, \bibinfo{person}{Jonathon
  Shlens}, {and} \bibinfo{person}{Zbigniew Wojna}.}
  \bibinfo{year}{2016}\natexlab{}.
\newblock \showarticletitle{Rethinking the Inception Architecture for Computer
  Vision}. In \bibinfo{booktitle}{\emph{IEEE/CVF Conference on Computer Vision
  and Pattern Recognition (CVPR)}}.
\newblock


\bibitem[\protect\citeauthoryear{Tao~Xu}{Tao~Xu}{2018}]%
        {attngan}
\bibfield{author}{\bibinfo{person}{Qiuyuan Huang Han Zhang Zhe Gan Xiaolei
  Huang Xiaodong~He Tao~Xu, Pengchuan~Zhang}.} \bibinfo{year}{2018}\natexlab{}.
\newblock \showarticletitle{AttnGAN: Fine-Grained Text to Image Generation with
  Attentional Generative Adversarial Networks}. In
  \bibinfo{booktitle}{\emph{IEEE/CVF Conference on Computer Vision and Pattern
  Recognition (CVPR)}}.
\newblock


\bibitem[\protect\citeauthoryear{Wah, Branson, Welinder, Perona, and
  Belongie}{Wah et~al\mbox{.}}{2011}]%
        {cub}
\bibfield{author}{\bibinfo{person}{C. Wah}, \bibinfo{person}{S. Branson},
  \bibinfo{person}{P. Welinder}, \bibinfo{person}{P. Perona}, {and}
  \bibinfo{person}{S. Belongie}.} \bibinfo{year}{2011}\natexlab{}.
\newblock \bibinfo{booktitle}{\emph{{The Caltech-UCSD Birds-200-2011
  Dataset}}}.
\newblock \bibinfo{type}{{T}echnical {R}eport} CNS-TR-2011-001.
  \bibinfo{institution}{California Institute of Technology}.
\newblock


\bibitem[\protect\citeauthoryear{{Wang}, {Yu}, {Dong}, and {Change Loy}}{{Wang}
  et~al\mbox{.}}{2018}]%
        {sftgan}
\bibfield{author}{\bibinfo{person}{X. {Wang}}, \bibinfo{person}{K. {Yu}},
  \bibinfo{person}{C. {Dong}}, {and} \bibinfo{person}{C. {Change Loy}}.}
  \bibinfo{year}{2018}\natexlab{}.
\newblock \showarticletitle{Recovering Realistic Texture in Image
  Super-Resolution by Deep Spatial Feature Transform}. In
  \bibinfo{booktitle}{\emph{IEEE/CVF Conference on Computer Vision and Pattern
  Recognition (CVPR)}}. \bibinfo{pages}{606--615}.
\newblock
\urldef\tempurl%
\url{https://doi.org/10.1109/CVPR.2018.00070}
\showDOI{\tempurl}


\bibitem[\protect\citeauthoryear{Wang, Yu, Wu, Gu, Liu, Dong, Loy, Qiao, and
  Tang}{Wang et~al\mbox{.}}{2018}]%
        {esrgan}
\bibfield{author}{\bibinfo{person}{Xintao Wang}, \bibinfo{person}{Ke Yu},
  \bibinfo{person}{Shixiang Wu}, \bibinfo{person}{Jinjin Gu},
  \bibinfo{person}{Yihao Liu}, \bibinfo{person}{Chao Dong},
  \bibinfo{person}{Chen~Change Loy}, \bibinfo{person}{Yu Qiao}, {and}
  \bibinfo{person}{Xiaoou Tang}.} \bibinfo{year}{2018}\natexlab{}.
\newblock \showarticletitle{{ESRGAN:} Enhanced Super-Resolution Generative
  Adversarial Networks}. In \bibinfo{booktitle}{\emph{European Conference on
  Computer Vision (ECCV)}}.
\newblock


\bibitem[\protect\citeauthoryear{Yang, Wang, Xie, Deng, and Tao}{Yang
  et~al\mbox{.}}{2021}]%
        {msaan}
\bibfield{author}{\bibinfo{person}{Yanhua Yang}, \bibinfo{person}{Lei Wang},
  \bibinfo{person}{De Xie}, \bibinfo{person}{Cheng Deng}, {and}
  \bibinfo{person}{Dacheng Tao}.} \bibinfo{year}{2021}\natexlab{}.
\newblock \showarticletitle{Multi-Sentence Auxiliary Adversarial Networks for
  Fine-Grained Text-to-Image Synthesis}.
\newblock \bibinfo{journal}{\emph{IEEE Transactions on Image Processing}}
  \bibinfo{volume}{30} (\bibinfo{year}{2021}), \bibinfo{pages}{2798--2809}.
\newblock
\urldef\tempurl%
\url{https://doi.org/10.1109/TIP.2021.3055062}
\showDOI{\tempurl}


\bibitem[\protect\citeauthoryear{Zhang, Koh, Baldridge, Lee, and Yang}{Zhang
  et~al\mbox{.}}{2021}]%
        {cmcl}
\bibfield{author}{\bibinfo{person}{Han Zhang}, \bibinfo{person}{Jing~Yu Koh},
  \bibinfo{person}{Jason Baldridge}, \bibinfo{person}{Honglak Lee}, {and}
  \bibinfo{person}{Yinfei Yang}.} \bibinfo{year}{2021}\natexlab{}.
\newblock \showarticletitle{Cross-Modal Contrastive Learning for Text-to-Image
  Generation}. In \bibinfo{booktitle}{\emph{IEEE/CVF Conference on Computer
  Vision and Pattern Recognition (CVPR)}}.
\newblock


\bibitem[\protect\citeauthoryear{Zhang, Xu, Li, Zhang, Wang, Huang, and
  Metaxas}{Zhang et~al\mbox{.}}{2017}]%
        {stackgan}
\bibfield{author}{\bibinfo{person}{Han Zhang}, \bibinfo{person}{Tao Xu},
  \bibinfo{person}{Hongsheng Li}, \bibinfo{person}{Shaoting Zhang},
  \bibinfo{person}{Xiaogang Wang}, \bibinfo{person}{Xiaolei Huang}, {and}
  \bibinfo{person}{Dimitris Metaxas}.} \bibinfo{year}{2017}\natexlab{}.
\newblock \showarticletitle{StackGAN: Text to Photo-realistic Image Synthesis
  with Stacked Generative Adversarial Networks}. In
  \bibinfo{booktitle}{\emph{IEEE International Conference on Computer Vision
  (ICCV)}}.
\newblock


\bibitem[\protect\citeauthoryear{{Zhou Wang}, {Bovik}, {Sheikh}, and
  {Simoncelli}}{{Zhou Wang} et~al\mbox{.}}{2004}]%
        {ssim}
\bibfield{author}{\bibinfo{person}{{Zhou Wang}}, \bibinfo{person}{A.~C.
  {Bovik}}, \bibinfo{person}{H.~R. {Sheikh}}, {and} \bibinfo{person}{E.~P.
  {Simoncelli}}.} \bibinfo{year}{2004}\natexlab{}.
\newblock \showarticletitle{Image quality assessment: from error visibility to
  structural similarity}.
\newblock \bibinfo{journal}{\emph{IEEE Transactions on Image Processing}}
  \bibinfo{volume}{13}, \bibinfo{number}{4} (\bibinfo{year}{2004}),
  \bibinfo{pages}{600--612}.
\newblock


\end{thebibliography}

\end{document}